\documentclass{article}

\usepackage{CJKutf8} 
\usepackage{longcat_style}
\usepackage{adjustbox}
\usepackage[utf8]{inputenc} %
\usepackage[T1]{fontenc}    %
\usepackage{newunicodechar}
\usepackage{hyperref}       %
\usepackage{xcolor}
\usepackage[normalem]{ulem} %
\hypersetup{
    colorlinks=true,      %
    linkcolor=blue,      %
    urlcolor=blue,       %
    citecolor=blue,      %
    linkbordercolor=blue, %
    urlbordercolor=blue,
    citebordercolor=blue,
    pdfborderstyle={/S/U/W 1}, %
}

\usepackage{float}
\usepackage{url}            %
\usepackage{booktabs}       %
\usepackage{amsfonts}       %
\usepackage{nicefrac}       %
\usepackage{microtype}      %
\usepackage{lipsum}		%
\usepackage{graphicx}
\usepackage{natbib}
\usepackage{doi}
\usepackage{amsmath}
\usepackage{amssymb} %
\usepackage{xspace}
\usepackage{enumitem}
\usepackage{multirow}
\usepackage{subcaption} 
\usepackage{makecell}
\usepackage{hyperref, cleveref}
\usepackage{pifont}
\usepackage[inkscapelatex=false]{svg}

\setlist[itemize]{leftmargin=*}
\setlist[enumerate]{leftmargin=*}
\setlist[description]{leftmargin=*}

\definecolor{midnightgreen}{rgb}{0.0, 0.29, 0.33}

\title{UNO-Bench: A Unified Benchmark for Exploring the Compositional Law Between Uni-modal and Omni-modal in Omni Models}

\author{
  Chen Chen\thanks{Equal contribution.} \quad
  Zeyang Hu\footnotemark[1] \quad
  Fengjiao Chen\thanks{Project Lead.} \\
  \bf{Liya Ma \quad Jiaxing Liu \quad Xiaoyu Li\footnotemark[2] \quad Ziwen Wang \quad Xuezhi Cao \quad Xunliang Cai} \\
  Meituan, Beijing, China. \\
  \texttt{\{chenchen165, huzeyang, chenfengjiao02, maliya06\}@meituan.com} \\
  \texttt{\{liujiaxing10, lixiaoyu28, wangziwen03, caoxuezhi, caixunliang\}@meituan.com}
}

\clearpage

\renewcommand{\headeright}{\raisebox{-0.2\height}{\includegraphics[height=1.8em]{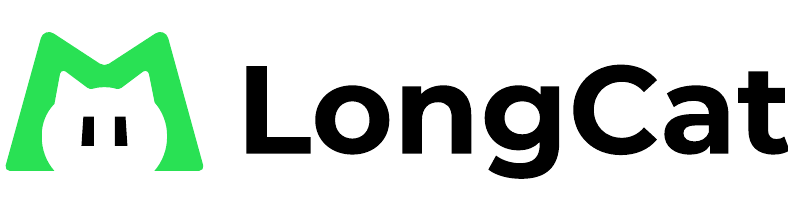}}}
\renewcommand{\shorttitle}{UNO-Bench: A Unified Benchmark for Omni Models}

\begin{document}
\maketitle

\begin{abstract}

Multimodal Large Languages models have been progressing from uni-modal understanding toward unifying visual, audio and language modalities, collectively termed omni models. However, the correlation between uni-modal and omni-modal remains unclear, which requires comprehensive evaluation to drive omni model's intelligence evolution. In this work, we introduce a novel, high-quality, and \textbf{UN}ified \textbf{O}mni model benchmark, \textbf{UNO-Bench}. This benchmark is designed to effectively evaluate both \textbf{UN}i-modal and \textbf{O}mni-modal capabilities under a unified ability taxonomy, spanning 44 task types and 5 modality combinations. It includes 1250 human curated samples for omni-modal with 98\% cross-modality solvability, and 2480 enhanced uni-modal samples. The human-generated dataset is well-suited to real-world scenarios, particularly within the Chinese context, whereas the automatically compressed dataset offers a 90\% increase in speed and maintains 98\% consistency across 18 public benchmarks. In addition to traditional multi-choice questions, we propose an innovative multi-step open-ended question format to assess complex reasoning. A general scoring model is incorporated, supporting 6 question types for automated evaluation with 95\% accuracy.
Experimental result shows the \textbf{Compositional Law} between omni-modal and uni-modal performance and the omni-modal capability manifests as a bottleneck effect on weak models, while exhibiting synergistic promotion on strong models.

\textbf{GitHub}: \href{https://github.com/meituan-longcat/UNO-Bench}{https://github.com/meituan-longcat/UNO-Bench} \\
\textbf{Hugging Face}: \href{https://huggingface.co/datasets/meituan-longcat/UNO-Bench}{https://huggingface.co/datasets/meituan-longcat/UNO-Bench}

\end{abstract}


\begin{figure}[h!]
    \centering
    \includegraphics[width=0.9\linewidth]{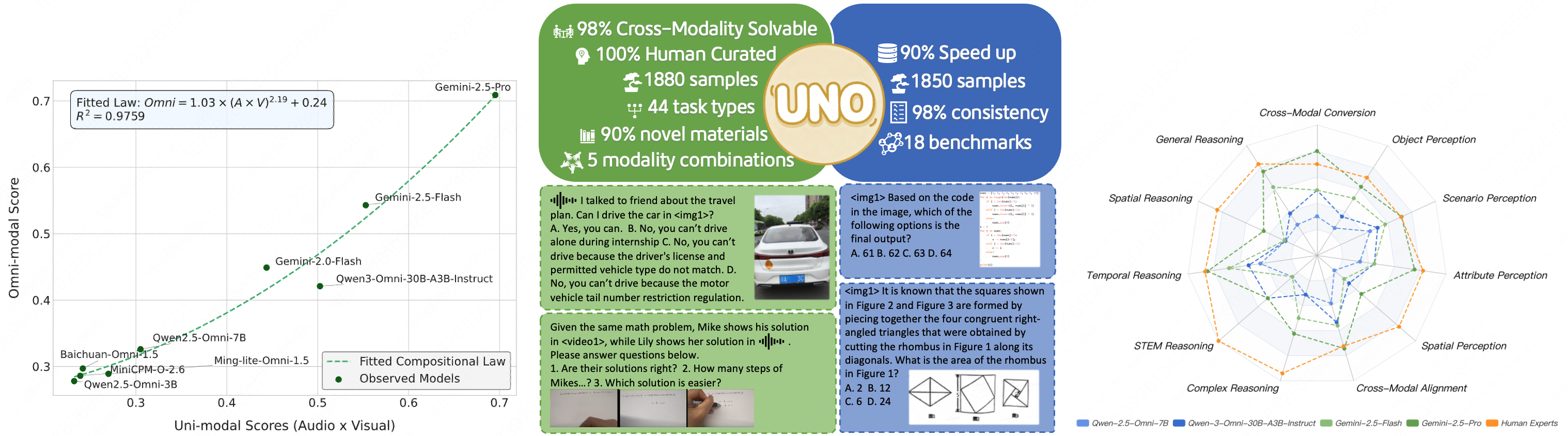}
    \caption{Benchmark Statistics and Evaluation Results.}    
    \label{fig:head_img}
\end{figure}

\clearpage
\tableofcontents
\clearpage

\section{Introduction}

Multimodal artificial intelligence has undergone extensive researches in visual language model and audio language model, with current advancements progressing toward unifying visual, audio and language modalities, collectively termed omni models. The evaluation paradigm for these models has consequently expanded from assessing uni-modal understanding capabilities (i.e. visual understanding, audio understanding) to the next-level of intelligence, omni-modal understanding. 

Existing omni model evaluation benchmarks remain relatively scarce and focus on different aspects. For example, some prioritize image comprehension\citep{omnibench}, others emphasize video understanding\citep{worldsense}, while a subset concentrates on speech interaction\citep{avodyssey}. Notably, existing datasets are exclusively English-centric, lacking evaluation benchmarks for Chinese linguistic contexts. 

The ideal omni model should simultaneously preserve visual understanding capabilities (e.g., MMBench\citep{mmbench}/MathVista\citep{mathvista}/MVBench\citep{li2024mvbench}), speech interaction proficiency (e.g., MMAU\citep{mmau}), and cross-modal integration capacity (e.g., OmniBench\citep{omnibench}/WorldSense\citep{worldsense}). Current evaluation paradigms employ disjointed benchmark suites for separate capability testing, creating resource-intensive evaluation processes and disconnected modality assessments. 
Beyond uni-modal, omni-modal capability introduces advanced challenges across image, video and audio modality. However, 77\% questions from WorldSense are solvable without vision or audio, and 25\% questions from OmniBench contain erroneous answers. 
These issues limit the evaluation and analysis of omni models' capabilities. 



Due to the limited quality and coverage of existing benchmarks, we introduce a novel and unified benchmark \textbf{UNO-Bench}. As shown in Figure.\ref{fig:data_ability}, the materials are collected from \textbf{human crafting} which prevents data contamination while better aligning with real-world scenarios. Beyond conventional multiple-choice questions, the evaluation adopts an innovative \textbf{Multi-Step Open-Ended Question} type to show a more realistic and discriminative evaluation result on complex reasoning. Besides the human crafted dataset, we incorporate existing uni-modal datasets by aggregating them systematically and design a \textbf{clustering-guided sampling method} to achieves both evaluation efficiency and consistency. In this way, our benchmark involves a comprehensive assessment necessitating omni models to maintain their uni-modal ability while simultaneously acquiring omni-modal capability.

\textbf{Main Contributions:}

1. Propose the first \textbf{UN}ified \textbf{O}mni model benchmark, \textbf{UNO-Bench}, which efficiently assesses both \textbf{UN}i-modal and \textbf{O}mni-modal understanding capabilities. UNO-Bench verifies the \textbf{Compositional Law} between omni-modal and uni-modal capability. The omni-modal capability acts as a bottleneck effect on weaker models, but shows synergistic enhancement on stronger models.

2. Establish a \textbf{high quality and diversity dataset construction pipeline} including human-centric process and automated data compression. As a result, UNO-Bench comprises 1250 human curated samples for omni-modal with \textbf{98\% cross-modality solvability}, and 2480 enhanced samples for uni-modal, \textbf{across 44 task types and 5 modality combinations}. The human-created novel dataset is well-suited to real-world scenarios, particularly within the Chinese context, whereas the automatically compressed dataset \textbf{offers a 90\% increase in speed and maintains 98\% consistency across 18 public benchmarks}.
Its comprehensive quality and efficiency significantly surpasses existing datasets.


3. Beyond conventional multiple-choice question type, the evaluation incorporates innovative \textbf{Multi-Step Open-Ended Question (MO)} to show a more realistic and discriminative evaluation result on complex reasoning especially for multi-step reasoning across modalities. For automated evaluation, a \textbf{General Scoring Model} is proposed to support 6 kinds of question types with 95\% accuracy on OOD models and benchmarks.

\begin{table*}
    \centering
    \resizebox{0.8\textwidth}{!}{  
      \begin{tabular}{l|c|c|c|c|c|c|c|c|c}
        \toprule
        \textbf{Dataset}     & \textbf{Omni-modal}  & \textbf{Uni-modal} & \textbf{Acc.} & \textbf{Solvable} & \textbf{Source} & \textbf{\#Tasks} & \textbf{\#QA Pairs} & \textbf{QA Type} & \textbf{Language}\\
        \midrule
        MMBench     & \ding{55}  & I        & -     & -  & 80\% private & 20  & 3217 & MC       & EN/CH \\
        MMAU        & \ding{55}  & A        & -     & -  & 15\% private & 27  & 10000 & MC       & EN \\
        MVBench     & \ding{55}  & V        & -     & -  & public      & 20  & 4000  & MC       & EN \\
        \midrule
        OmniBench   & I+A       & \ding{55} & 75\%  & 90\%   & public      & 8   & 1142 & MC       & EN    \\
        AV-Odyssey  & I+V+A     & \ding{55} & 91\%  & 99\%   & public      & 26  & 4555 & MC       & EN    \\
        WorldSense  & V+A       & \ding{55} & 99\%  & 23\%   & public      & 26  & 3172 & MC       & EN    \\
        Daily-Omni  & V+A       & \ding{55} & 94\%& 59\%& public      & 6   & 1197 & MC       & EN \\
        \midrule
        \textbf{UNO-Bench\small{-omni}} & I+V+A  & -  & ~100\% & 98\% & 90\% private & 44& 1250 & MC/MO & EN/CH \\
        \textbf{UNO-Bench\small{-uni}} & -  & I/V/A  & - & - & 40\% private      & 44&  2480     & MC       &EN/CH \\
        \bottomrule
      \end{tabular}
    }
    \caption{Comparison of MultiModal Benchmarks, with I, A, V, and T representing image, audio, video, and text modalities, respectively. It reports on the accuracy of question-answer pairs and the percentage of questions requiring omni-modal solutions, labeled as Acc. and Solvable. The Source category specifies the origin of the materials. Private sources, as opposed to public ones, can prevent data contamination. QA types include MC for multi-choice questions and MO for multi-step open-ended questions. EN and CH denote English and Chinese languages. UNO-Bench includes 1250 human-curated samples in the omni-modal section (referred to as -omni) and 2480 enhanced samples in the uni-modal section (referred to as -uni).}
    \label{tab:statistic}
\end{table*}

\section{Related Work}
\subsection{Uni-Modal Benchmarks}
Based on large language models, vision language models (VLM) \citep{Qwen2.5-VL, coreteam2025mimovltechnicalreport, zeng2025glm} and audio language models (ALM) \citep{ding2025kimi, wu2025step} introduce the general intelligence to vision modality and audio modality respectively. Various uni-modal benchmarks conduct comprehensive evaluations on VLMs \citep{mmbench, mathvista, wang2024mathvision, wang2024charxiv, liu2024ocrbench, mathew2021docvqa, ouyang2024omnidocbench, li2024mvbench, wu2024longvideobench, liu2024tempcompass, realworldqa, xiao2021nextqa, huang2025ovbench, hu2025videommmu, fu2024videomme} and ALMs \citep{ardila2019common, wang2021covost, yang2024air, ao2024sd}. For image modality, MMBench\citep{mmbench} proposed a systematically designed benchmark to evaluate general image understanding on 20 different tasks. Focused on mathematic, MathVision\citep{wang2024mathvision} collected questions from 19 mathematic competitions to evaluate VLMs complex reasoning ability. In addition to above, OCRBench\citep{liu2024ocrbench} supplied the evaluation on text recognition and document understanding. For video modality, MVBench\citep{li2024mvbench} aggregated 11 public video benchmarks and incorporated data enhancement process to cover 20 dynamic video understanding tasks. To complement the long video understanding area, LongVideoBench\citep{wu2024longvideobench} introduces hourly video materials to evaluate the information retrieval ability from long context. For audio modality, MMAU\citep{mmau} provides general audio understanding assessment across speech, sounds and music domains, featuring diverse audio samples. There are massive uni-modal benchmarks covering diverse model abilities on vision modality and audio modality separately.

\subsection{Omni-Modal Benchmarks}
\begin{figure*}
    \centering
    \includegraphics[width=\linewidth]{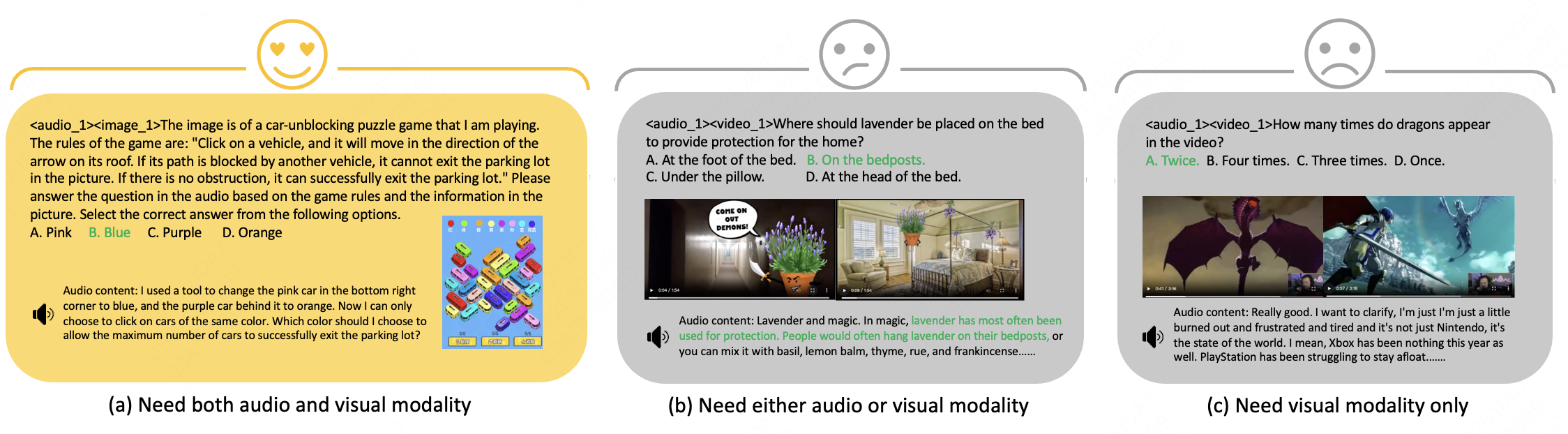}
    \caption{Illustration of the cross-modality solvable sample.}
    \label{fig:modality_solvable}
\end{figure*}

Omni models have arisen in recent years\citep{comanici2025gemini, qwen3omni, mingomni, li2025baichuanomni15technicalreport}, as the pioneer, Gemini\citep{comanici2025gemini} shows a strong ability in understanding both vision and audio, while Qwen-3-Omni\citep{qwen3omni} provides leading performance in open-source models. However, there are less omni-modal benchmarks that can evaluate the modality combination across image, video and audio. OmniBench\citep{omnibench} inserted audio as a context into the image understanding task and made up an omni-modal benchmark, while the data quality needs further improvement. WorldSence\citep{worldsense} emphasized audio-visual data in real world scenarios with high data quality, while most audio-visual questions can be solved by audio or video alone, which cannot assess the cross-modality ability. Other datasets focus on audio \citep{avodyssey} or video \citep{zhou2025dailyomni} and cover limited task types.
For instance, in Figure.\ref{fig:modality_solvable}(b), the problem can be resolved using either the audio modality or the visual modality, whereas in Figure.\ref{fig:modality_solvable}(c), only the visual modality is necessary to address the problem. These instances are likely to exaggerate the capabilities of the omni model, making it crucial to evaluate the cross-modality solvable problem (illustrated in Figure.\ref{fig:modality_solvable}(a)) to accurately assess omni-modal capability (refer to Section.\ref{sec:composition_law} for more details).

Addressing these limitations, we propose a novel and unified benchmark, UNO-Bench, that enables comprehensive model assessment and pushing omni model to the next-level of intelligence. 



\section{Method}
In this section, we first introduce the omni-modal dataset construction pipeline in Section.\ref{sec:data_pipeline}. For uni-modal dataset, an quality improvement method and a general dataset compression method to improve the evaluation efficiency are introduced in Section.\ref{sec:uni-modal data_pipeline}. Finally, the novel multi-step open-ended questions are presented alongside a general scoring model in Section.\ref{Sec:Multiple Open-ended Questions}

\subsection{Omni-modal Dataset Construction}
\label{sec:data_pipeline}


We have established a human-centric data construction pipeline (Figure.\ref{fig:data_pipeline}) that efficiently empower human intelligence to produce high-quality and high-diversity dataset.


\begin{figure}[h!]
    \centering
    \includegraphics[width=0.9\linewidth]{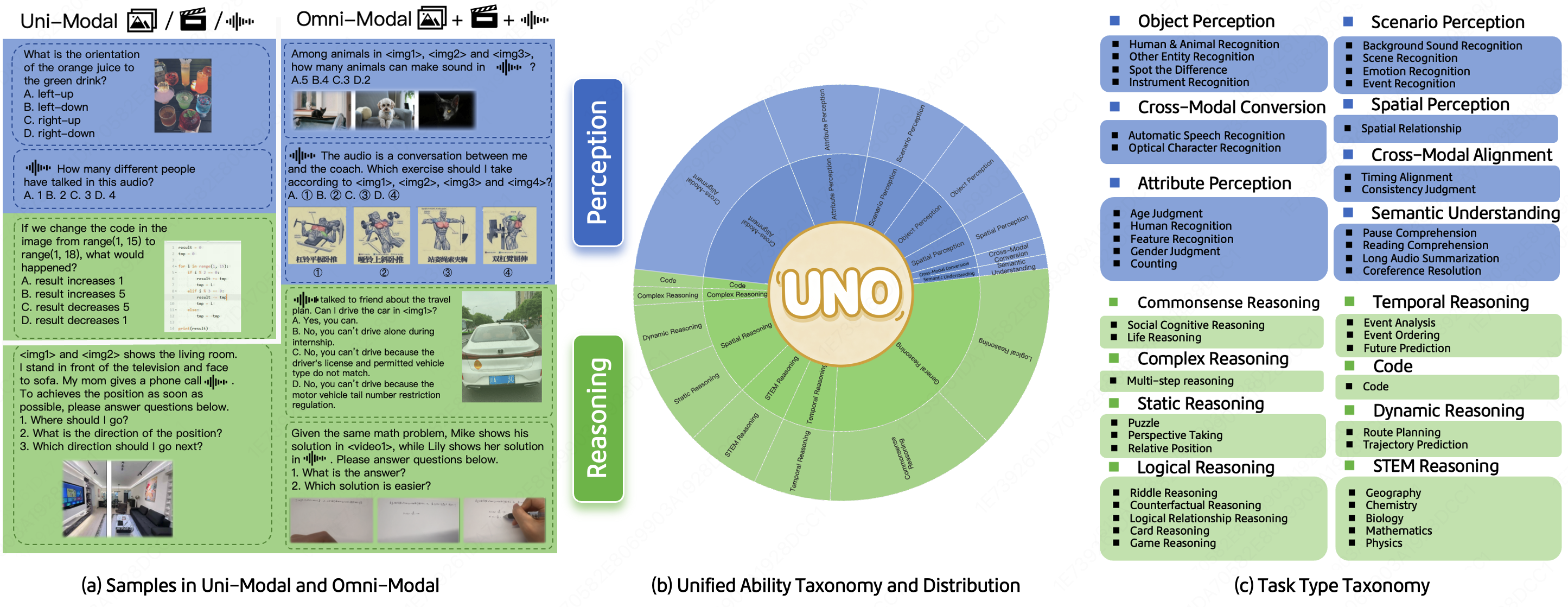}
    \caption{Illustration of the unified ability taxonomy proposed in UNO-Bench.}
    \label{fig:data_ability}
\end{figure}

\subsubsection{Model Ability Taxonomy}
\label{sec:model_ability_sys}

Through cumulative experiences on multimodal evaluation from both model-side and user-side, we summarize the capabilities of uni-modal and omni-modal into a unified model ability taxonomy. 
As shown in Figure.\ref{fig:data_ability}(b), the omni model's capabilities are systematically categorized into two primary dimensions: Perception and Reasoning. Detailed definitions and examples can be found in the Appendix.\ref{sec:appendix_model ability}.

\textbf{Perception} dimension structured through seven recognition types including 
Object Perception,
Attribute Perception,
Scenario Perception,
Spatial Perception,
Cross-Modal Conversion,
Semantic Understanding.
In addition, we incorporate Cross-Modal Alignment to assess information synchronization across modalities. 

\textbf{Reasoning} dimension extends conventional reasoning categories (including General, STEM, Code) with Spatial Reasoning (including Static Reasoning and Dynamic Reasoning), Temporal Reasoning, and Complex Reasoning (which indicates multi-conditional, multi-step problem).

As shown in Figure.\ref{fig:data_ability}(a), the unified ability taxonomy combines uni-modal and omni-modal abilities which provides a comprehensive measurement that is particularly critical for omni models. For example, Scenario Perception includes the recognition of visual scenes and the judgment of audio scenes. Based on this taxonomy, we create a diversity dataset with 44 task types illustrated in Figure.\ref{fig:data_ability}(c).

\subsubsection{Material Collection}
In both data quality checks and experimental results, we found that the natural video with audio-visual synchronized data contains a large amount of information redundancy, only a few videos require both audio and visual modality simultaneously. Therefore, we begin with carefully designed material collection.

Our materials have the following three characteristics: 

\textbf{Diverse Sources}. The majority of our materials are real-world photos and videos collected through crowd sourcing, and another portion sourced from copyright-free websites. Additionally, a small fraction comes from high-quality public datasets such as MMVU\citep{zhao2025mmvu}, LongVideoBench\citep{wu2024longvideobench}, and VideoVista\citep{chen2025videovista}. 

\textbf{Rich and Diverse Topics}. Our materials cover a broad spectrum of subjects, including society, culture, art, life, literature, science, and so on. 

\textbf{Live-Recorded Audio}. Apart from background sounds and music, all dialogue is recorded by human speakers. With over 20 participants in the recording process, the audio features are rich and closely reflect the diverse vocal characteristics of the real world, such as Mandarin and Sichuan dialect.

Finally, we conduct material filtering. Eliminate meaningless, illogical, and low-quality materials, and categorize the remaining materials by theme to create a material library. Additionally, label the materials with more detailed information such as subject, event, scene, and style to facilitate subsequent annotators to quickly find matching materials.

\subsubsection{QA Annotation}
Our annotators consist of human experts and high-quality crowd-sourced users. Human experts have extensive experience in cross-modal data construction and annotation, a deeper understanding of model capabilities, and thus ensure higher professionalism and specificity in the data they construct. Most crowd-sourced users are college students with rich experience in multimodal model interactions and diverse professional backgrounds, providing data with better authenticity and diversity. 

\textit{First}, annotators clarify the required image/video features based on task type definitions and filter appropriate materials from existing libraries using tags. \textit{Second}, following data construction requirements, they then design prompts and corresponding answers. \textit{Third}, to enhance data authenticity, all dialogue audio is recorded manually. Through this workflow, we ultimately generate complete QA pairs encompassing three modalities: visual, auditory, and textual.

Compared to conventional methods limited to human intervention only during the quality assurance phase, our pipeline integrates a \textbf{human-centric} approach, ensuring continuous manual involvement from the initial data sourcing to the final output. This methodology not only prevents data leakage but also more accurately simulates real-world scenarios. Furthermore, the manually curated Chinese dataset genuinely captures user requirements in a Chinese linguistic context, compensating for the shortcomings of most existing English-centric datasets.

\subsubsection{Quality Inspection}
To ensure the data quality, we have established a multi-stage, cyclically validated quality assurance system composed of automated tools and manual review. Each question undergoes at least three rounds of independent quality inspection to maximize data quality. \textbf{Model Check}, a preliminary model check is conducted to filter out cases with ambiguous questions, non-unique answers, or those that do not conform to the task type. \textbf{Ablation Study}, through modality ablation experiments, we remove one modality of information from the QA pair to see if the model can answer based solely on the remaining information. If the question becomes unsolvable or ambiguous after removing any one modality, it proves the cross-modality solvability of the data. \textbf{Human Check}, finally manual quality inspection and revision are performed.

\begin{figure}[h!]
    \centering
    \includegraphics[width=0.9\linewidth]{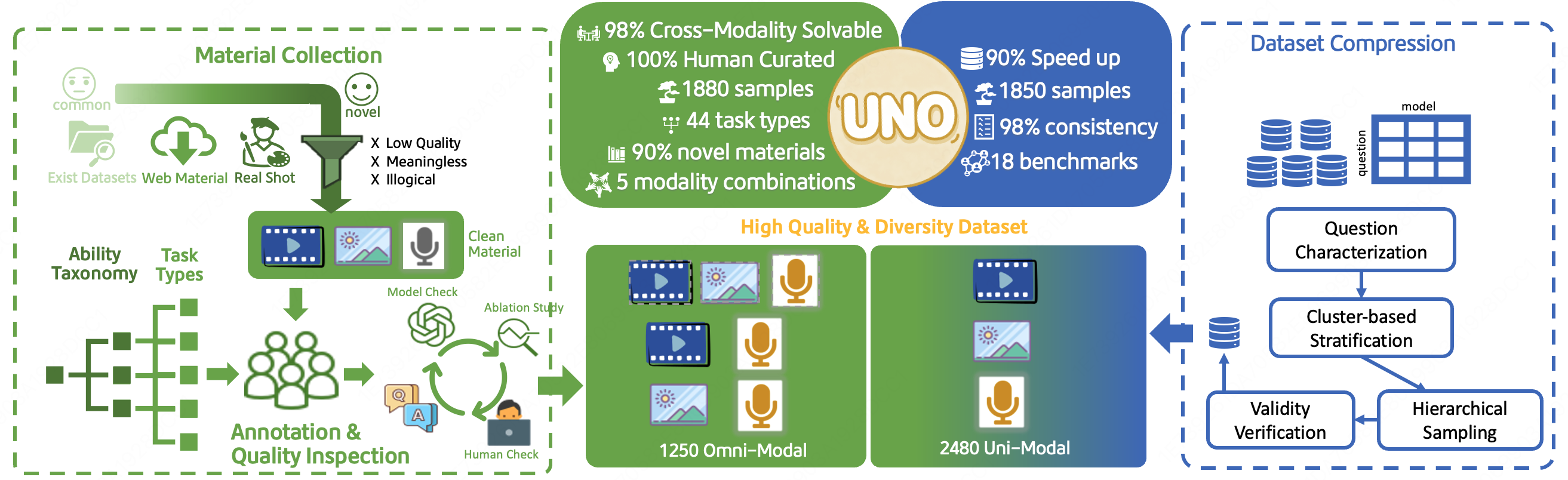}
    \caption{Dataset Construction Pipeline includes human-centric process (left side) and automated data compression (right side). First, we collect diverse and novel materials to prevent data contamination. Second, with the proposed unified ability taxonomy, human annotators including experts will craft questions, answers and record audios in real-world scenarios. Finally, with model checking, ablation study and human experts revision, we achieves high quality and diversity dataset. Regarding automated data compression, we present a clustering-guided hierarchical sampling method to achieve efficient compression while maintaining high evaluation consistency.}    
    \label{fig:data_pipeline}
\end{figure}

\subsection{Uni-modal Dataset Improvement}
\label{sec:uni-modal data_pipeline}
\subsubsection{Quality Improvement}
Existing public uni-modal datasets are bothered by data leakage issue\citep{xu2024leakage}. To verify the influence, we adopt privatization improvement on the widely used public dataset MMBench\citep{mmbench}. As shown in Figure.\ref{fig:mmbench_private}, the performance of models have better distinguishability after dataset improvement, reflecting the true capability differences between models. Therefore, for uni-modal data, we also follow the aforementioned construction process for self-construction datasets.
In addition to self-constructed data, we also selected some multimodal data from public datasets to supplement in terms of capability items and data types. (Data mainly comes from AV-Odyssey\citep{avodyssey} and WorldSense\citep{worldsense}, accounting for 11\% of the total). The specific selection logic is as follows:

\textbf{Comprehensiveness}: In terms of capabilities, focus on supplementing the perception part with a relatively low self-construction proportion, while also adding some reasoning questions; in terms of data types, prioritize selecting the video plus audio modality combination with a lower self-construction proportion for supplementation, followed by image plus audio.

\textbf{Diversity}: Supplement material types not covered in self-construction data to enhance diversity.

\textbf{High Quality}: Pay attention to the quality of datasets (whether uni-modal answers are reasonable and accurate).

\textbf{Discriminative}: Pay attention to the performance of this dataset on the model, and remove overly difficult subsets with little discrimination.

\subsubsection{Dataset Compression}
\label{sec:data_compresssion}
Regarding the existing large-scale uni-modal benchmarks, to reduce the evaluation cost of large-scale models, we designed a \textbf{clustering-guided hierarchical sampling (CGHS)} method as shown in Figure.\ref{fig:data_pipeline}. CGHS is a general method for dataset compression, which utilizes model performance metrics as features rather than the content of questions to select important samples that impact model performance. For training datasets, CGHS can retain both simple and difficult samples in an unsupervised manner or minimize similar rollout samples in a batch for online policy. When it comes to test datasets, CGHS is capable of achieving efficient compression while maintaining high evaluation consistency. The introduction of CGHS is outlined in the following steps:

\textbf{Question Characterization}: Represent each question as an x-dimensional vector, where dimensions correspond to scores from different models on that question.

\textbf{Cluster-based Stratification}: Utilize the Kmeans++\citep{arthur2007k} algorithm to categorize questions into k clusters, each representing a "model performance similar" question type (e.g., easy questions, difficult questions, etc.).

\textbf{Hierarchical Sampling}: Determine the sample size for each stratum based on cluster size proportions, and construct the final evaluation subset through simple random sampling.

\textbf{Validity Verification}: To verify the compression performance, we define these metrics: Spearman's Rank Correlation Coefficient (SRCC) for ranking consistency, Pearson's Linear Correlation Coefficient (PLCC) for linear value consistency, Root Mean Square Error (RMSE) for numerical precision, Margin of Error (MoE) for quantifying estimation uncertainty, and Confidence Interval Coverage (CIC) for statistical reliability.

To ensure statistical stability, we repeat the above steps by using 5 random splits and performing 10-fold cross-validation. This approach identifies the optimal sample size via cost-benefit curve analysis, leading to a reduction in evaluation costs by over 90\% while preserving accuracy, as shown in Figure.\ref{fig:data_compression}. 

\subsection{Multi-Step Open-Ended Questions}
\label{Sec:Multiple Open-ended Questions}

\subsubsection{Question Type Definition}

Evaluating the multi-step reasoning capabilities of omni models presents a significant challenge. Real-world problems require models to integrate multi-modal information and execute a sequence of logical steps. However, current automated benchmarks, often relying on Outcome Reward Models (ORMs), typically provide only a binary pass/fail judgment. This approach fails to distinguish between a model that completes 80\% of a task and one that fails at 20\%, a crucial gap that human evaluators easily perceive. While alternatives like Process Reward Models (PRMs)\cite{lightman2023let} or multi-turn dialogues\cite{reddy2019coqa} exist, they are hampered by high implementation difficulty, low accuracy, or poor efficiency. Moreover, the prevalence of multiple-choice formats in existing benchmarks is unrepresentative of real-world, open-ended user queries and may conceal the weaknesses of models.

To address these issues, we propose an innovative Multi-Step Open-Ended Question (MO) type, designed for granular and realistic assessment. In the construction of MO dataset, complex problems are first deconstructed by human experts into a series of progressive, interdependent sub-questions. Each sub-question is assigned a score based on its importance, summing to a total of 10 points. During testing, all sub-questions are posed in a single turn, requiring the model to generate a step-by-step open-ended response. 
This method allows us to precisely quantify how far along a complex reasoning chain a model can proceed, offering a more accurate and insightful measure of its true capabilities. An example is shown in Figure.\ref{fig:multi_qa_demo}.

\subsubsection{General Scoring Model}
\begin{figure}[t]
  \centering
  \begin{minipage}[b]{0.54\textwidth}
    \centering
    \includegraphics[width=\columnwidth]{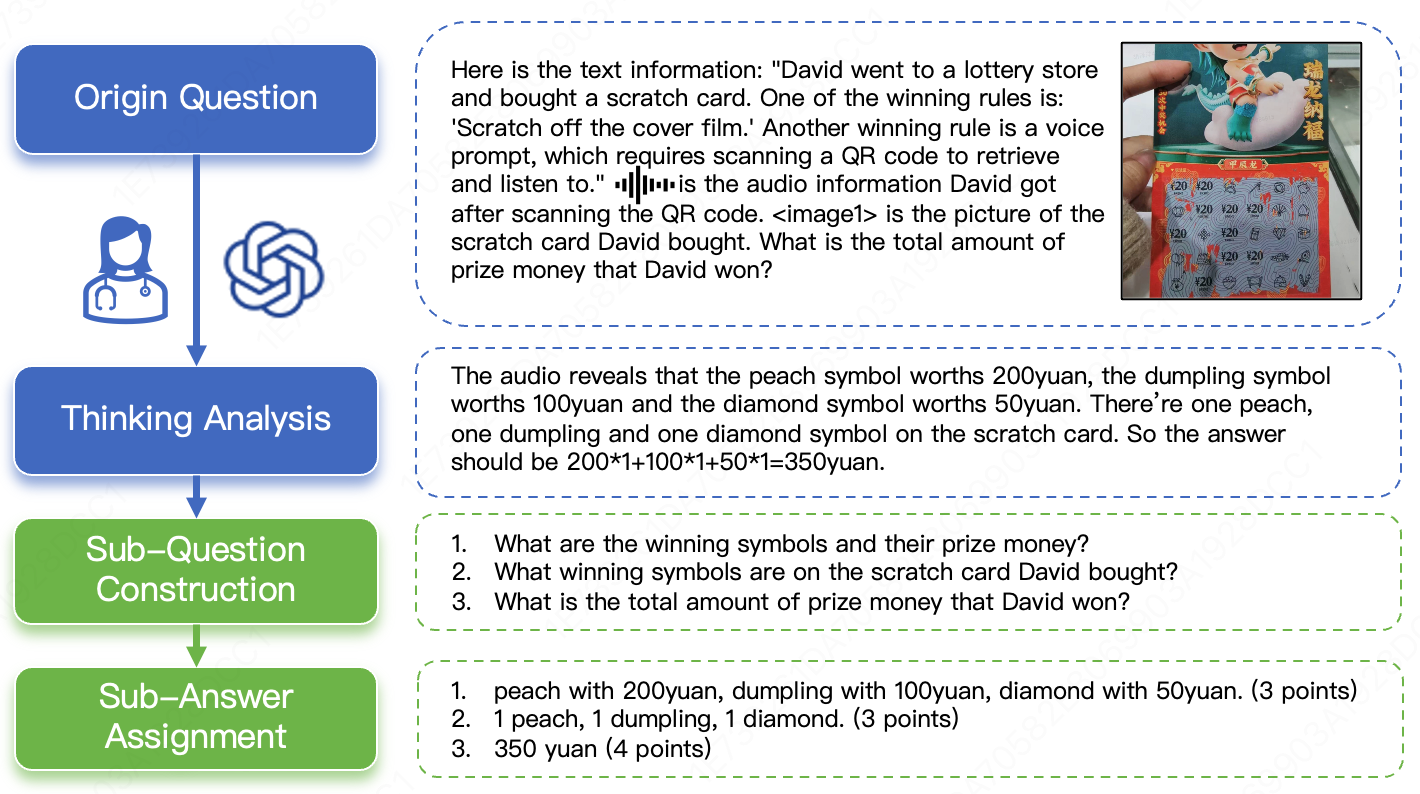}
    \caption{Construction of Multi-Step Open-Ended Questions. }
    \label{fig:multi_qa_demo}
  \end{minipage} 
  \hfill
  \begin{minipage}[b]{0.45\textwidth}
    \centering
    \includegraphics[width=\columnwidth]{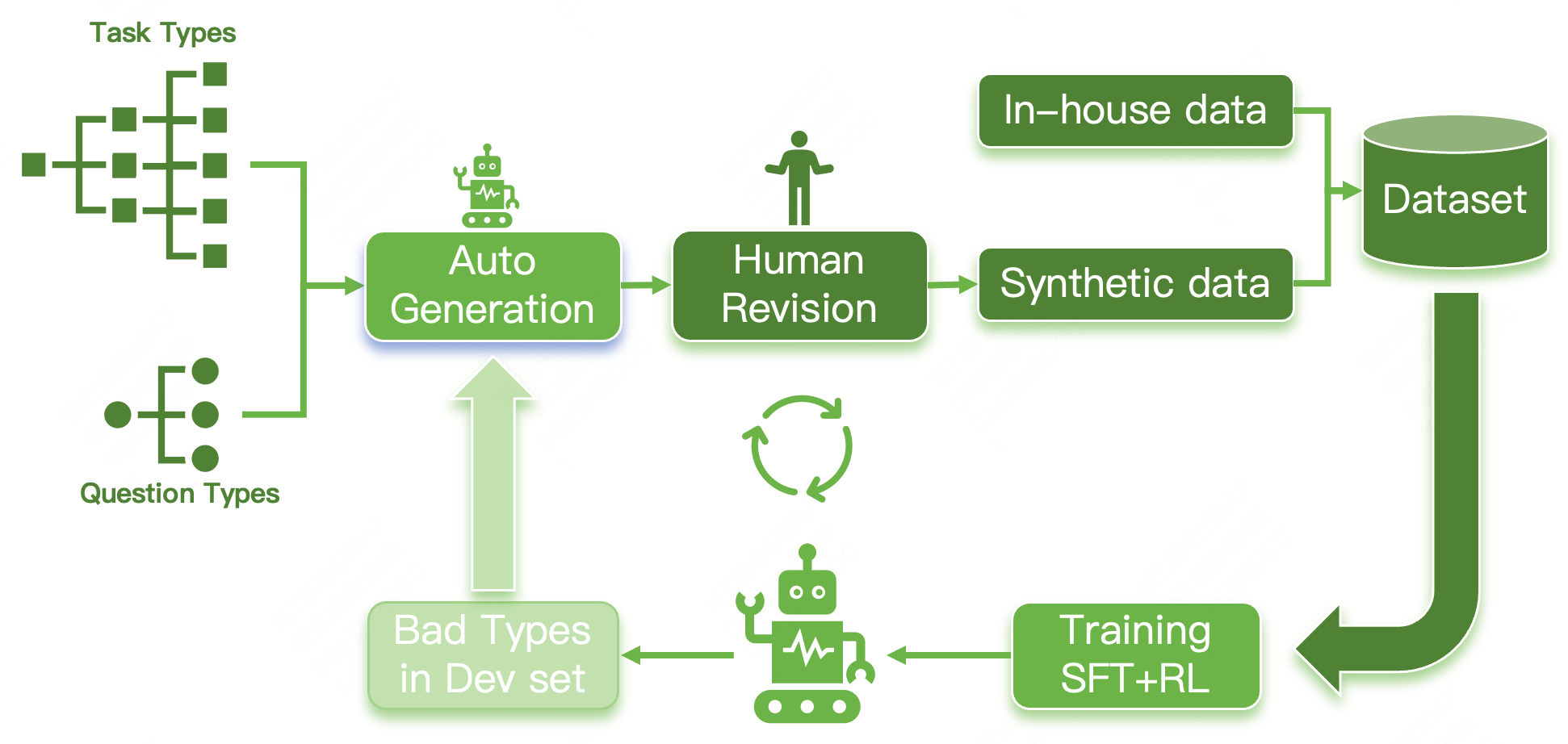}
    \caption{Training pipeline for general scoring model.}
    \label{fig:score_model_pipeline}
  \end{minipage}
\end{figure}

\begin{figure*}
    \centering
    \includegraphics[width=0.9\linewidth]{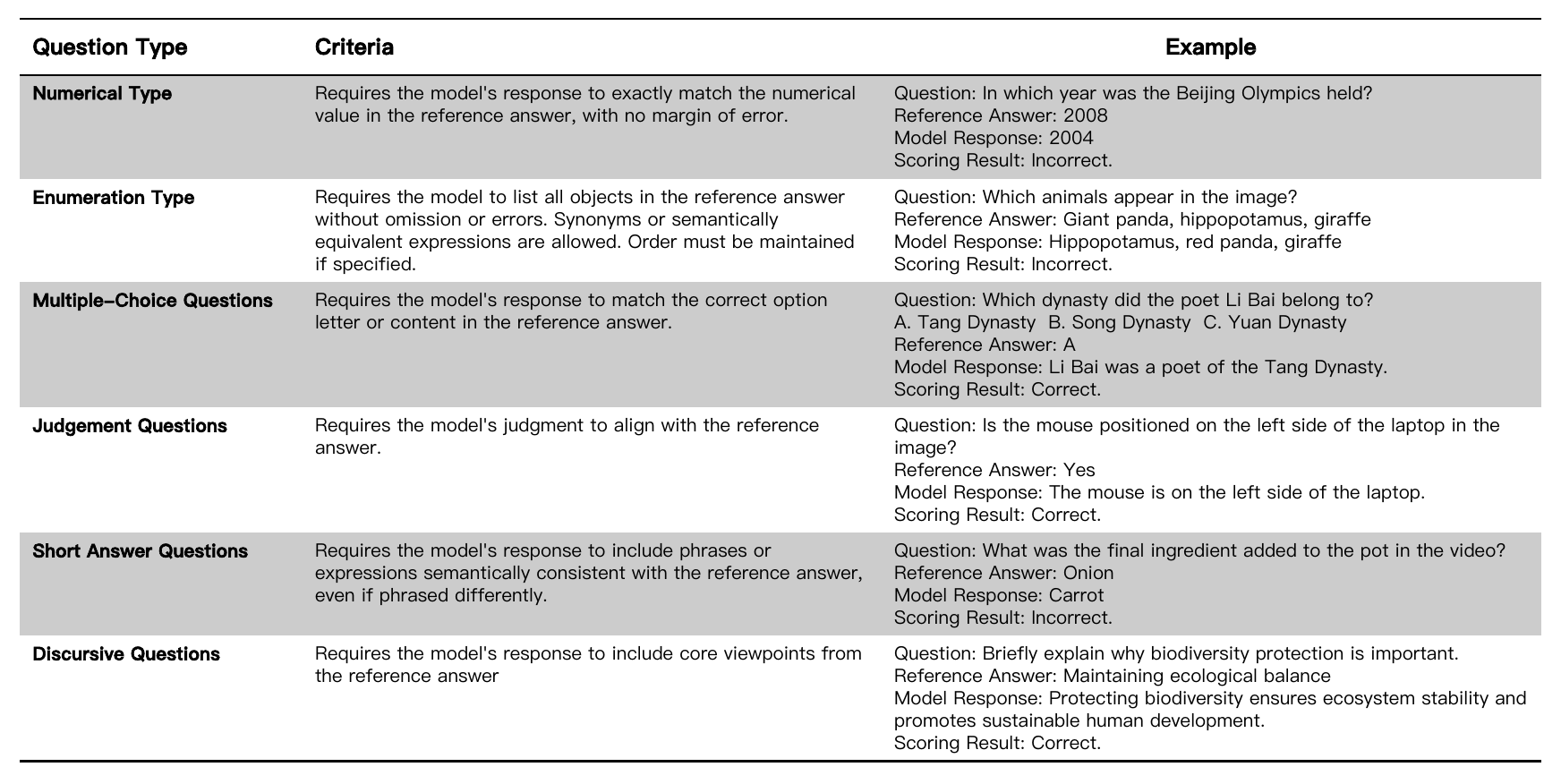}
    \caption{Definition of finer question types for general scoring model.}
    \label{fig:score_questin_type}
\end{figure*}

Beside the dataset construction, multi-step open-ended question introduces a new challenge of automated evaluation. To overcome this obstacle, we propose a general scoring model that supports multi-choice question, single-step open-ended question and multi-step open-ended question at the same time. Since the task is to compare the target answer and the predicted answer, we use Qwen3-14B\citep{qwen3} as backbone and curate a training dataset as illustrated in Figure.\ref{fig:score_model_pipeline}. One of the critical way to improve accuracy is to group questions into finer types and define appropriate criteria for each types, as shown in Figure.\ref{fig:score_questin_type}.
Through the human-in-the-loop dataset curation, the scoring model achieves 95\% accuracy in out-of-distribution models and benchmarks.

Experiments in Section.\ref{sec:exp_multi_qa} show that compared with single-step evaluation method (e.g. multiple-choice questions), multi-step open-ended questions can effectively observe the ability decay of models in long-chain reasoning, providing a more realistic difficulty for advanced models with stronger discrimination.

\section{Experiment and Analysis}
\subsection{Experiment Setting}
We evaluate omni models that support text, visual, and audio inputs simultaneously, including open-source models: Qwen-3-Omni-30B-A3B-Instruct\citep{qwen3omni}, Qwen-2.5-Omni-3B, Qwen-2.5-Omni-7B\citep{xu2025qwen25omnitechnicalreport}, Baichuan-Omni-1.5\citep{li2025baichuanomni15technicalreport}, MiniCPM-O-2.6\citep{yao2024minicpm}, and Ming-lite-Omni-1.5\citep{mingomni}, as well as closed-source models: Gemini-2.5-Pro, Gemini-2.5-Flash, and Gemini-2.0-Flash\citep{comanici2025gemini}. To have a fair comparison between instruct model and thinking model, we adopt similar way in Qwen-3\citep{qwen3omni} that limits thinking budget to 128 tokens. We apply this restriction to Gemini-2.5-Pro and disable the thinking mode for both Gemini-2.5-Flash and Gemini-2.0-Flash. All the other model integrations strictly adhere to official implementations. In video processing, each model receives raw video and performs frame sampling according to its own sampling strategy.

\newpage

In the subsequent sections, we perform detailed experiments on UNO-Bench and aim to address the following questions:

\begin{enumerate}
\item How do current omni models perform, and what are their limitations?
\item How are uni-modal and omni-modal capabilities related?
\item Is the UNO-Bench capable of effectively evaluating the omni model?
\end{enumerate}
\subsection{Model Performance}

\subsubsection{Overall Analysis}

Our main evaluation, summarized in Table.\ref{tab:eval_overall}, reveals a clear performance hierarchy where proprietary models, particularly Gemini-2.5-Pro, establish the state-of-the-art across all benchmarks. Meanwhile, progress within the open-source community is notable, with increased model scale and more training data, exemplified by Qwen-3-Omni-30B, leading to substantial improvements. Furthermore, we observe a strong positive correlation between a model's performance on the foundational Audio and Visual tasks and its scores on the more demanding Omni benchmarks, suggesting that robust uni-modal perception is a prerequisite for advanced omni-modal understanding.

On the \textbf{Omni-MC} (Multiple-Choice) benchmark, which evaluates omni-modal comprehension, smaller open-source models exhibit performance marginally surpassing the random guess baseline (25.00), achieving scores between 27.80 and 29.70. The larger Qwen-3-Omni-30B marks a significant leap, with a score of 42.10 that approaches the performance of entry-level proprietary models like Gemini-2.0-Flash (44.90). Nevertheless, a substantial performance deficit persists when compared to the leading Gemini-2.5-Pro (70.90). This gap highlights the profound difficulty of advanced omni-modal comprehension, even in a multiple-choice format.

The \textbf{Omni-MO} (Multi-Step Open-Ended) benchmark presents a considerably greater challenge, as evidenced by the universal and marked degradation in performance for all models relative to their Omni-MC scores. This degradation reveals a systemic limitation in multi-step omni-modal reasoning. For instance, the leading model, Gemini-2.5-Pro, attained a score of merely 57.32 on this benchmark, reflecting a decline of 13.58 points relative to its performance on the Omni-MC task. In comparison, the highest-scoring open-source model, Qwen-3-Omni-30B, achieved only 37.08 points.

To dissect the models' core capabilities, we perform a fine-grained analysis based on our proposed ability taxonomy, with detailed results presented in Table.\ref{tab:eval_ability}.

In perception, a notable trend emerges: while smaller models find Recognition easier than Alignment, more powerful models like Qwen-3-Omni-30B-A3B and the Gemini-2.5 series exhibit stronger Alignment capabilities. This suggests that advanced models develop a more sophisticated grasp of inter-modal relationships. Among open-source models, Qwen-3-Omni-30B-A3B achieves the highest perception score (49.02). Gemini-2.5-Pro significantly leads overall, with both its Alignment (74.35) and Recognition (70.05) scores surpassing 70.

In reasoning, Spatial Reasoning is consistently the most challenging task across all models. Even the top-performing Gemini-2.5-Pro only achieves 45.00. Notably, Baichuan-Omni-1.5 demonstrates the best spatial reasoning among open-source models with a score of 28.33. For General and Temporal Reasoning, the new Qwen-3-Omni-30B-A3B establishes itself as the open-source leader.

Overall, reasoning proves to be a more challenging frontier than perception. This is highlighted by the performance gap between the leading proprietary model, Gemini-2.5-Pro, and the best open-source model, Qwen-3-Omni-30B-A3B. The disparity is 23.04 points in Perception (72.06 vs. 49.02) but widens to a more substantial 33.00 points in Reasoning (70.41 vs. 37.41). This indicates that advanced reasoning remains a key differentiator and a primary bottleneck for current multimodal models.

\begin{table*}[htbp]
    \caption{General performance of omni models in UNO-Bench for both uni-modal capability and omni-modal capability, where omni-modal benchmark includes multi-choice questions (Omni-MC) and multi-step open-ended questions (Omni-MO).}
    \label{tab:eval_overall} 
    \centering
    \footnotesize 
    \begin{tabular}{l|c|c|c|c}
        \toprule
        Model & Audio & Visual & Omni-MC & Omni-MO \\
        \midrule
        Qwen-2.5-Omni-3B& 54.40 & 42.67 & 27.80 & 24.76 \\
        MiniCPM-O-2.6                   & 56.50 & 42.27 & 28.60 & 23.76 \\
        Ming-lite-Omni-1.5              & 58.30 & 46.28 & 28.90 & 25.48 \\
        Baichuan-Omni-1.5               & 54.10 & 44.66 & 29.70 & 21.04 \\
        Qwen-2.5-Omni-7B& 60.20 & 50.68 & 32.60 & 27.72 \\
        Qwen-3-Omni-30B-A3B& \textbf{79.40} &\textbf{ 63.29} & \textbf{42.10} & \textbf{37.08} \\
        \midrule
        Gemini-2.0-Flash                & 70.70 & 62.76 & 44.90 & 38.56 \\
        Gemini-2.5-Flash    & 79.50 & 69.54 & 54.30 & 47.08 \\
        Gemini-2.5-Pro     & \textbf{88.40} & \textbf{78.67} & \textbf{70.90} & \textbf{57.32} \\
        \bottomrule
    \end{tabular}
\end{table*}

\begin{table*}[htbp]
    \caption{Analysis of Omni-MC on ability taxonomy. To simplify the analysis, Cross-modal Recognition refers to the set of other Perception capabilities except Cross-modal Alignment.}
    \centering
    \resizebox{0.9\textwidth}{!}{
        \begin{tabular}{l|ccc|cccc|c}
            \toprule
            \multirow{3}{*}{Model} & \multicolumn{3}{c|}{Perception} & \multicolumn{4}{c|}{Reasoning} & \multirow{3}{*}{Overall} \\
            \cline{2-8}
            & \begin{tabular}[c]{@{}c@{}}Cross-modal \\ Alignment\end{tabular} & \begin{tabular}[c]{@{}c@{}}Cross-modal \\ Recognition\end{tabular} & Overall & \begin{tabular}[c]{@{}c@{}}General \\ Reasoning\end{tabular} & \begin{tabular}[c]{@{}c@{}}Temporal\\ Reasoning\end{tabular} & \begin{tabular}[c]{@{}c@{}}Spatial \\ Reasoning\end{tabular} & Overall \\
            \midrule
            Qwen-2.5-Omni-3B            & 29.84 & 35.94 & 33.09 & 20.65 & 50.00 & 20.83 & 23.98 & 27.80 \\
            MiniCPM-O-2.6               & 26.70 & 30.88 & 28.92 & 26.62 & 42.42 & 26.67 & 28.40 & 28.60 \\
            Ming-lite-Omni-1.5          & 28.80 & 35.94 & 32.60 & 24.38 & 43.94 & 24.17 & 26.53 & 28.90 \\
            Baichuan-Omni-1.5           & 30.89 & 32.26 & 31.62 & 25.87 & 45.45 & 28.33 & 28.57 & 29.70 \\
            Qwen-2.5-Omni-7B            & 38.22 & 36.41 & 37.25 & 28.11 & 43.94 & 26.67 & 29.59 & 32.60 \\
            Qwen-3-Omni-30B-A3B & \textbf{53.40} & \textbf{45.16} & \textbf{49.02} & \textbf{38.06} &\textbf{53.03} & \textbf{26.67} & \textbf{37.41} & \textbf{42.10} \\
            \midrule
            Gemini-2.0-Flash            & 43.98 & 49.77 & 47.06 & 45.02 & 57.58 & 31.67 & 43.71 & 44.90 \\
            Gemini-2.5-Flash            & 56.02 & 50.69 & 53.19 & 61.44 & 68.18 & 27.50 & 55.27 & 54.30 \\
            Gemini-2.5-Pro              & \textbf{74.35} & \textbf{70.05} & \textbf{72.06} & \textbf{75.62} & \textbf{84.85} & \textbf{45.00} & \textbf{70.41} & \textbf{70.90} \\
            \bottomrule
        \end{tabular}
    }
    \label{tab:eval_ability}
\end{table*}

\subsubsection{Top-tier Analysis}
\begin{figure}[t]
  \centering
  \begin{minipage}[b]{0.44\textwidth}
    \centering
    \includegraphics[width=\linewidth]{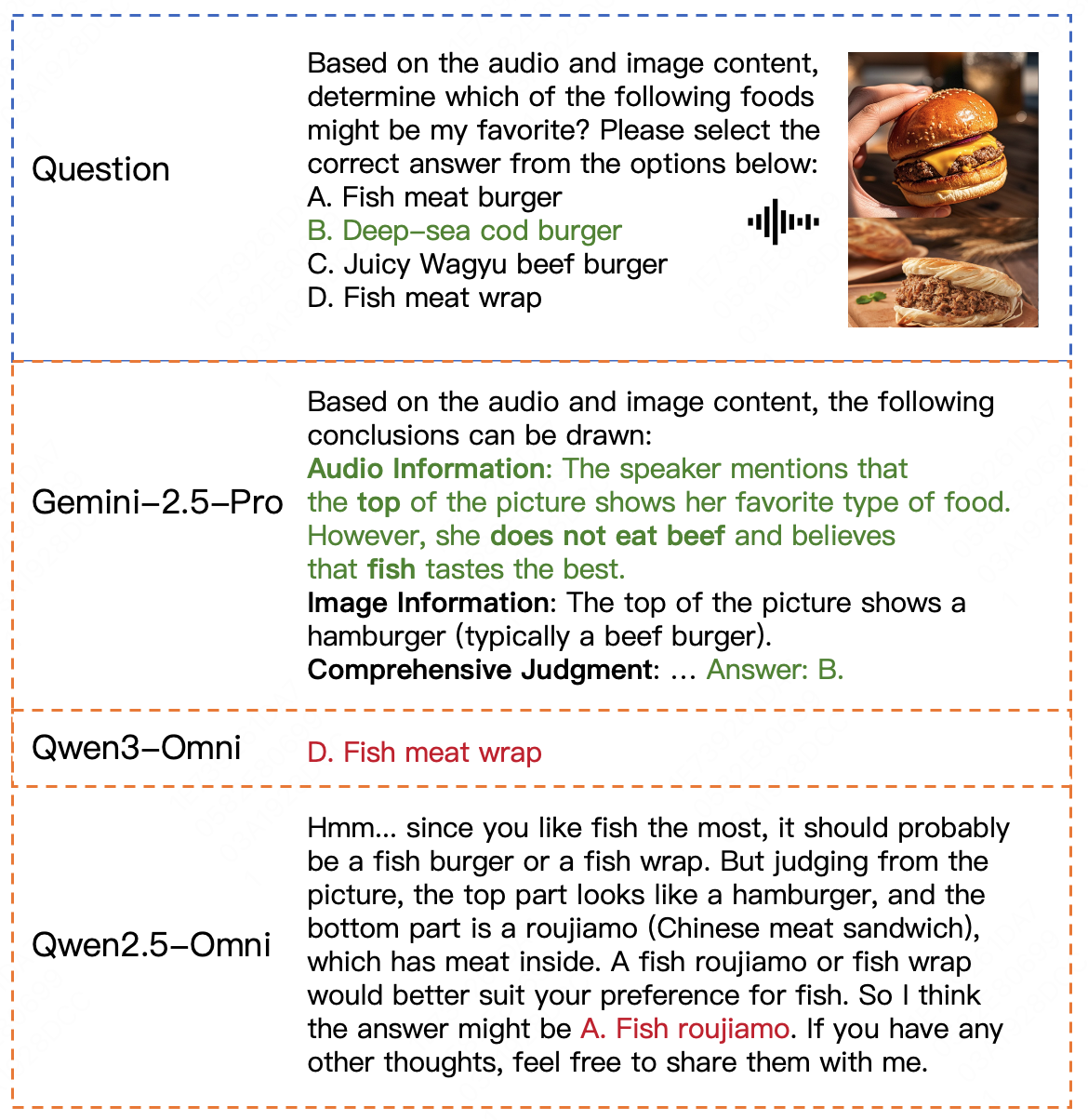}
    \caption{An example of omni-modal evaluation result. Gemini-2.5-Pro displays audio captions to assist in solving the problem.}
    \label{fig:audio_caption}
  \end{minipage} 
  \hfill
  \begin{minipage}[b]{0.55\textwidth}
    \centering
    \includegraphics[width=\linewidth]{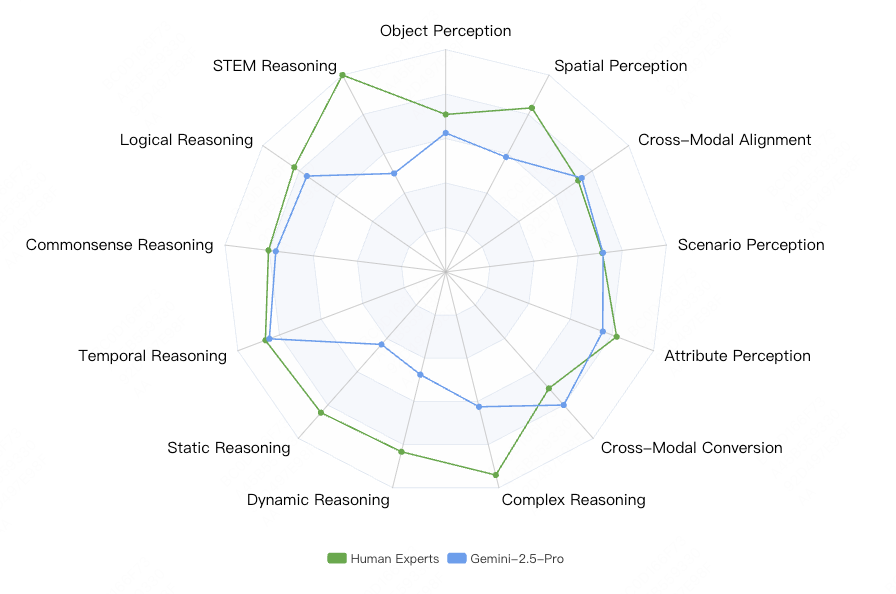}
    \caption{The competition between human experts and Gemini-2.5-Pro. Gemini-2.5-Pro shows comparable perception capability but lower reasoning capability.}
    \label{fig:Human VS Gemini-2.5-Pro}
  \end{minipage}
\end{figure}

What makes the performance of Gemini-2.5-Pro stand out compared to other models? We aim to offer an analysis along with several hypotheses.
On one hand, it stems from the leading uni-modal understanding ability. On the other hand, regarding the reasoning mechanism, Gemini is equipped with audio captioning functionalities as indicated in the technical report\citep{comanici2025gemini}, and illustrated in Figure.\ref{fig:audio_caption}. It can also naturally incorporate audio content as a foundation for reasoning. Existing open-source models, due to their smaller size, lack reasoning processes in a multimodal context. Limited reasoning mostly relies on text or images, with few involving specific audio content.

The successive question is whether Gemini's performance measures up. To answer this question, we invited human experts for a competition. It's crucial to highlight that, unlike the dataset annotators, these experts had no prior exposure to the questions or answers.

\textbf{Finding 1. Gemini-2.5-Pro has reached human comparable perception ability in omni-modal perception, yet there remains a gap in its reasoning performance.}
Compared to human experts, Gemini-2.5-Pro exhibits similar performance in perception, but falls behind in reasoning. The comparison of scores for specific ability items can be seen in Figure.\ref{fig:Human VS Gemini-2.5-Pro}. Upon examining ability analysis, we observe an intriguing phenomenon: humans are more proficient in reasoning as opposed to perception (81.3\% compared to 74.3\%), which contrasts with the model's performance. By interviewing various human experts, it becomes evident that humans might miss some information in video or audio formats, and their world knowledge is more limited compared to large language models.


\subsection{Uni-Modal v.s. Cross-Modal}
\label{sec:composition_law}


To investigate the relationship of uni-modal and omni-modal understanding ability, we conduct regression analysis and ablation experiments. Thanks to the unified ability taxonomy and the high quality of omni-modal samples in UNO-Bench, we find some interesting observations.

\textbf{Finding 2. Compositional Law: the effectiveness of omni-modal capability is related to the product of the performances of individual modalities by a power-law.} 
Observing the results in Table.\ref{tab:eval_overall}, we identify a strong correlation between a model's omni-modal performance and its uni-modal capabilities. 
To formalize this, we derive a Compositional Law from a general functional form by applying two simplifying principles dictated by the omni-modal tasks proposed in our UNO-Bench. Let's elaborate on the specifics below.

\textbf{General Model \& Task Constraints.}
We begin by positing that the omni-modal performance $\mathcal{P}_{\text{Omni}}$ is a function of uni-modal performances $\mathcal{P}_{\text{A}}$ and $\mathcal{P}_{\text{V}}$. 
A general form can be written as:
\begin{equation}
    \begin{split}
        \mathcal{P}_{\text{Omni}}(\mathcal{P}_{\text{A}}, \mathcal{P}_{\text{V}}) ={}& f_{\text{A}}(\mathcal{P}_{\text{A}}) + f_{\text{V}}(\mathcal{P}_{\text{V}}) \\
        & + f_{\text{I}}(\mathcal{P}_{\text{A}}, \mathcal{P}_{\text{V}}) + b
    \end{split}
    \label{eq:general_form}
\end{equation}
where $f_{\text{A}}, f_{\text{V}}$ represent modality independent path contributions, $f_{\text{I}}$ the interaction, and $b$ a baseline performance constant (e.g. random guess).

We arrive at the following result through rigorous mathematical derivation, and the detailed derivation process is provided in the Appendix.\ref{sec:appendix_derivation_of_law}.
\begin{equation}
    \mathcal{P}_{\text{Omni}} = C \cdot \mathcal{P}_{\text{A}}^{\alpha} \mathcal{P}_{\text{V}}^{\beta} + b
    \label{eq:compositional_law_cobb}
\end{equation}
where $C$ is a scaling constant, and exponents $\alpha, \beta$ model the interaction's elasticity.

We then posit a \textbf{fusion symmetry} assumption: in end-to-end omni models, the fusion mechanism does not inherently favor one modality over another \citep{qwen3omni, yao2024minicpm}, implying symmetric scaling behavior. This leads to $\alpha = \beta$. Substituting this into Eq.~\ref{eq:compositional_law_cobb}, we arrive at the \textbf{Omni-modal Compositional Law}:
\begin{equation}
    \mathcal{P}_{\text{Omni}} = C \cdot (\mathcal{P}_{\text{A}} \times \mathcal{P}_{\text{V}})^{\alpha} + b
    \label{eq:final_law}
\end{equation}

where $\alpha$ is the synergistic exponent, $C$ is a scaling coefficient, and $b$ is a baseline bias.

\begin{figure}[t]
    \centering
    \includegraphics[width=0.9\columnwidth]{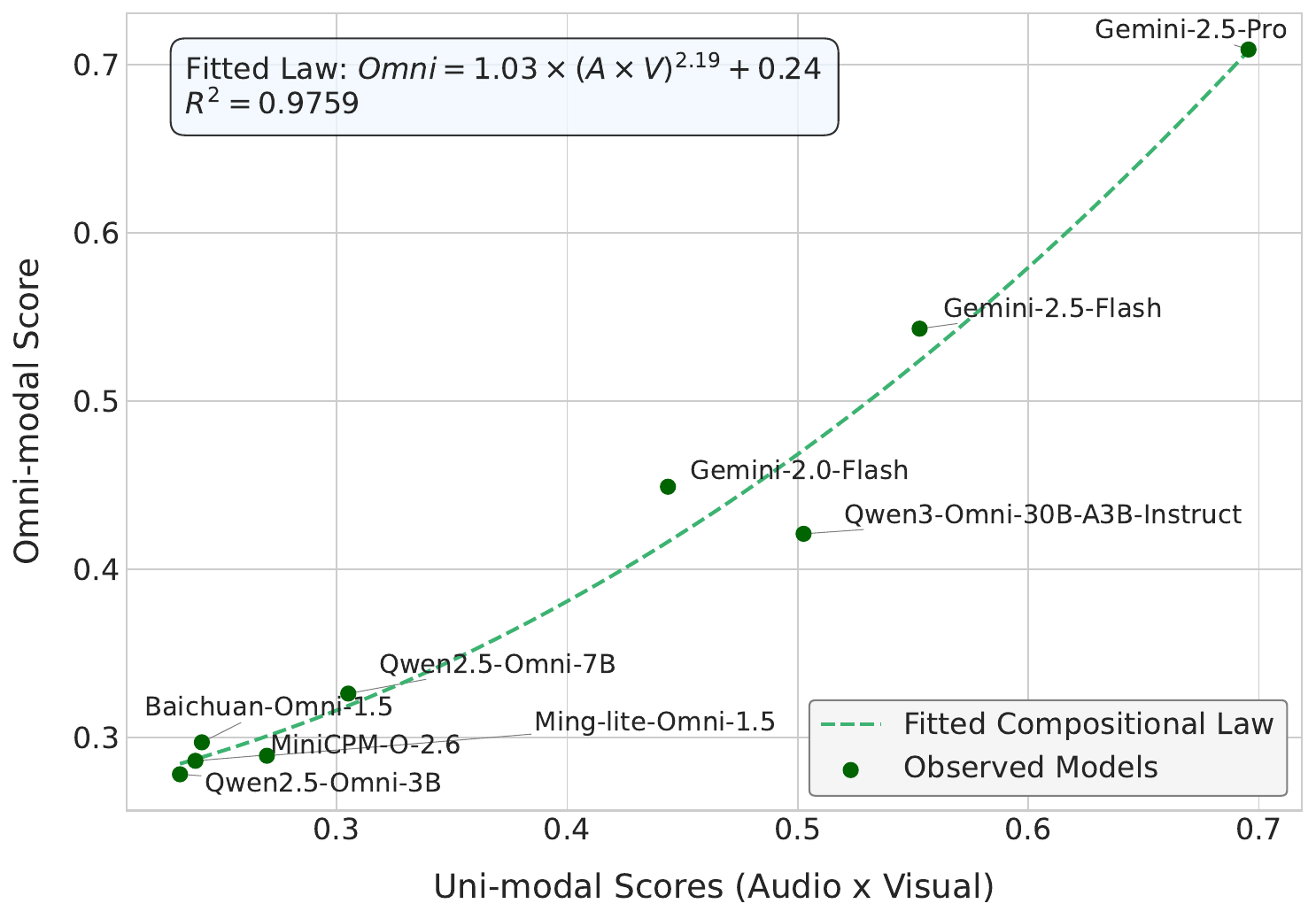} 
    \caption{\textbf{The Compositional Law of Omni-modal Performance.} Observed omni-modal scores (dots) versus the product of their uni-modal scores. The dashed line represents our fitted law (Eq.~\ref{eq:composition_law_fitted}), which shows a remarkable alignment with the empirical data ($R^2=0.9759$). The convex, accelerating curve visualizes the Power-law synergy.}
    \label{fig:compositional_law}
\end{figure}

A non-linear regression on data from leading models (Figure.\ref{fig:compositional_law}) yields the precise empirical formula:
\begin{equation}
    \mathcal{P}_{\text{Omni}} \approx 1.0332 \cdot (\mathcal{P}_{\text{A}} \times \mathcal{P}_{\text{V}})^{2.1918} + 0.2422
    \label{eq:composition_law_fitted}
\end{equation}
This model demonstrates an exceptional fit, with a coefficient of determination ($R^2$) of \textbf{0.9759}. Analysis of the fitted parameters reveals a clear transition from limited gains to emergent capabilities, driven by the super-linear nature of the law.

\paragraph{Power-law Synergy and Emergent Ability.} The exponent $\alpha \approx 2.19$ is the most critical discovery, revealing a powerful \textbf{Power-law synergy}. Because $\alpha > 1$, the function is convex, meaning the rate of performance gain accelerates. This explains the transition from a "short-board effect" to an "emergent ability" seen in Figure.\ref{fig:compositional_law}:
\begin{itemize}
    \item \textbf{Limited Gains at Low Performance:} For models with weaker uni-modal abilities (e.g., MiniCPM-O), the curve is relatively flat. Small improvements in the product of uni-modal scores yield only marginal gains in omni-modal performance. This can be seen as a "short-board effect", where the system is not yet capable of effectively leveraging the combined inputs.
    \item \textbf{Emergent Ability at High Performance:} As uni-modal abilities strengthen (e.g., moving towards Gemini-2.5-Pro), the curve steepens dramatically. The same amount of improvement in the uni-modal product now yields a much larger increase in omni-modal performance. This accelerating return on investment is the quantitative signature of \textbf{emergence}, where stronger foundational skills unlock disproportionately powerful combined capabilities.
\end{itemize}

\paragraph{Interpreting the Coefficients and Benchmark Coherence.} The other parameters complete the picture. The bias term $b \approx 0.2422$ acts as a performance floor. As uni-modal performances approach zero, the system's output converges to this value, which is strikingly close to the 0.25 random-guess accuracy of our benchmark. The scaling coefficient $C \approx 1.0332$, being remarkably close to unity, indicates a harmonious and naturally scaled system. We attribute this harmony not only to the models' intrinsic fusion mechanisms but also to the coherent design of our benchmark itself.

Additional fitted models are presented in Appendix.\ref{appendix:model_selection}. We argue that our proposed model is the most natural and interpretable among them. Interestingly, most models indicate that the visual modality tends to offer greater benefits than the audio modality. This phenomenon may be attributable to the relatively weaker visual capabilities of models at the current stage of development.

It is worth emphasizing that this finding is directly attributed to the deliberate design of UNO-Bench. Specifically, it not only ensures a balanced distribution of capabilities across both uni-modal and omni-modal tasks, but also constructs the majority of questions to demand the joint processing of both modalities for resolution.

Next, we conduct ablation studies to dig dive about the enhancement from vision and audio modality respectively.


\subsubsection{Ablation Visual Understanding}
To quantify the contribution of visual information, we conducted an ablation study with three settings: audio-only (Audio), audio plus high-quality textual captions of the visual scene (+ Caption), and the full audio-visual input (+ Visual). The captions were generated by Gemini-2.5-Pro to ensure descriptive richness. Results are detailed in Table.\ref{tab:eval_ablation_visual}.

With only audio input, most models' performance drops to a level near random guessing (around 20-28\%), confirming the critical role of visual context. A notable exception is Gemini-2.5-Pro, which scores 40.34, suggesting an ability to leverage linguistic cues or shortcuts within the questions even without visual data.

The introduction of \texttt{Caption} information yields significant but highly variable performance gains. Powerful models like the Gemini series and Qwen-3-Omni-30B-A3B demonstrate a substantial leap in performance (gains of 20-25 points), showcasing their strong ability to reconstruct scenes from textual descriptions. In contrast, models like MiniCPM-O-2.6 and Ming-lite-Omni-1.5 show minimal improvement, indicating a weaker capacity for this text-to-vision reasoning.

Comparing \texttt{Caption} against full \texttt{Visual} input reveals a fascinating dichotomy. For the most capable model, Gemini-2.5-Pro, direct visual information provides a clear advantage over captions (70.90 vs. 65.10), proving that raw visual data contains nuances that text cannot fully capture. However, for several other models, including Gemini-2.0-Flash and the powerful Qwen-3-Omni-30B-A3B, performance with captions is surprisingly on par with, or even slightly exceeds, that with direct visual input. This suggests that for these models, the language processing pathway may be more adept at extracting semantic meaning than their own visual encoders, highlighting a potential imbalance in their multimodal processing capabilities.

\begin{table*}[htbp]
    \caption{Ablation of visual understanding ability. The three settings are audio-only (\texttt{Audio}), audio plus high-quality textual captions of the visual scene (\texttt{+Caption}), and the full audio-visual input (\texttt{+Visual}). }
    \centering
    \resizebox{0.9\textwidth}{!}{
        \begin{tabular}{l|ccc|ccc|ccc}
            \toprule
            \multirow{2}{*}{Model} & \multicolumn{3}{c|}{Perception} & \multicolumn{3}{c|}{Reasoning} & \multicolumn{3}{c}{Overall} \\
            \cline{2-10}
            & Audio & + Caption & + Visual & Audio & + Caption & + Visual & Audio & + Caption & + Visual \\
            \midrule
            Qwen-2.5-Omni-3B            & 17.76 & 29.13 & 33.09 & 20.07 & 21.43 & 23.98 & 19.12 & 24.60 & 27.80 \\
            MiniCPM-O-2.6               & \textbf{29.44} & 29.61 & 28.92 & \textbf{27.21} & 29.93 & 28.40 & \textbf{28.13} & 29.80 & 28.60 \\
            Ming-lite-Omni-1.5          & 26.28 & 31.07 & 32.60 & 23.13 & 21.43 & 26.53 & 24.42 & 25.40 & 28.90 \\
            Baichuan-Omni-1.5           & 22.14 & 32.04 & 31.62 & 23.81 & 26.70 & 28.57 & 23.12 & 28.90 & 29.70 \\
            Qwen-2.5-Omni-7B            & 22.14 & 30.10 & 37.25 & 20.41 & 25.34 & 29.59 & 21.12 & 27.30 & 32.60 \\
            Qwen-3-Omni-30B-A3B         & 27.01 & \textbf{46.84} & \textbf{49.02} & 18.71 & \textbf{39.63} & \textbf{37.41} & 22.12 & \textbf{42.60} & \textbf{42.10} \\
            \midrule
            Gemini-2.0-Flash            & 25.55 & 44.17 & 47.06 & 29.76 & 45.58 & 43.71 & 28.03 & 45.00 & 44.90 \\
            Gemini-2.5-Flash            & 22.63 & 49.03 & 53.19 & 29.08 & 53.23 & 55.27 & 26.43 & 51.50 & 54.30 \\
            Gemini-2.5-Pro              & \textbf{37.71} & \textbf{63.83} & \textbf{72.06} & \textbf{42.18} & \textbf{65.99} & \textbf{70.41} & \textbf{40.34} & \textbf{65.10} & \textbf{70.90} \\
            \bottomrule
        \end{tabular}
    }
    \label{tab:eval_ablation_visual}
\end{table*}

\subsubsection{Ablation Audio Understanding}

To isolate the impact of auditory information, we evaluated models under three conditions: visual-only (\texttt{Visual}), visual plus transcribed audio (\texttt{+Caption}), and the full audio-visual input (\texttt{+Audio}). We further divided the audio into three categories: the \texttt{Speech} category was annotated with ASR transcripts, while both the \texttt{Environment} and \texttt{Music} categories received textual descriptions. To ensure the robustness of our analysis and improve statistical reliability, the data-insufficient \texttt{Music} class was merged with the \texttt{Environment} class. The majority of the transcriptions were manually produced by human annotators, while a smaller subset was generated by a powerful multimodal model. The results are presented in Table.\ref{tab:eval_ablation_audio_part2}.

The \texttt{Visual}-only setting results in significantly lower performance across all models, with Overall scores ranging from 21.20 to 33.70. This confirms the critical role of auditory context in multimodal understanding. The introduction of textual audio descriptions (\texttt{+Caption}) substantially boosts performance across the board. The improvement is particularly dramatic for high-capacity models like Gemini-2.5-Pro (+31.0 points Overall) and Qwen-3-Omni-30B-A3B (+17.4 points Overall), demonstrating their strong ability to integrate textual information.

The comparison between \texttt{+Caption} and \texttt{+Audio} reveals crucial insights into the models' raw audio processing capabilities. In environmental sound scenarios, understanding raw audio remains a significant challenge for most open-source models. For instance, Qwen-2.5-Omni-3B, MiniCPM-O-2.6, and Ming-lite-Omni-1.5 all exhibit considerably higher performance with textual descriptions (\texttt{+Caption}) than with the original audio (\texttt{+Audio}). This suggests that their audio encoders struggle to extract meaningful features from complex non-speech sounds, making them prefer clean textual summaries. In contrast, the most capable models—Gemini-2.5-Pro, Gemini-2.5-Flash, and Qwen-3-Omni-30B-A3B—demonstrate superior audio understanding by scoring higher in the \texttt{+Audio} setting, indicating they can extract richer information directly from the audio signal than is present in the provided caption.

In conversational (Speech) scenarios, the results are more nuanced. The top-performing Gemini-2.5-Pro shows a substantial advantage with raw audio over ASR transcripts (\texttt{+Audio} 72.16 vs. \texttt{+Caption} 66.00), indicating it effectively leverages paralinguistic cues such as tone, emotion, and prosody that are lost in transcription. Conversely, several other models, including the Qwen series and MiniCPM-O-2.6, perform slightly better with ASR transcripts (\texttt{+Caption}) than with raw audio. This points to a common bottleneck where imperfections in their audio encoders are a greater liability than the information lost during ASR, making clean text a more reliable input. Notably, Gemini-2.5-Flash achieves nearly identical scores in both settings, suggesting its ASR and audio understanding capabilities are exceptionally well-aligned.


\begin{table*}[htbp]
    \caption{Ablation of audio understanding ability. The three settings are visual-only (\texttt{Visual}), visual plus transcribed audio (\texttt{+Caption}), and the full audio-visual input (\texttt{+Audio}). We further divided the audio into two categories: Environment sounds, for which we provided textual descriptions, and Speech, for which we provided ASR transcripts.}
    \centering
    \resizebox{0.9\textwidth}{!}{
        \begin{tabular}{l|ccc|ccc|ccc}
            \toprule
            \multirow{2}{*}{Model} & \multicolumn{3}{c|}{Environment} & \multicolumn{3}{c|}{Speech} & \multicolumn{3}{c}{Overall} \\
            \cline{2-10}
            & Visual & +Caption & +Audio & Visual & +Caption & +Audio & Visual & +Caption & +Audio \\
            \midrule
            Qwen-2.5-Omni-3B      & 26.28 & 41.03 & 34.62 & 24.76 & 26.66 & 26.54 & 25.00 & 28.90 & 27.80 \\
            MiniCPM-O-2.6         & 26.92 & 39.74 & 34.62 & \textbf{28.08} & 28.44 & 27.49 & \textbf{27.90} & 30.20 & 28.60 \\
            Ming-lite-Omni-1.5    & 31.41 & 43.59 & 35.26 & 22.27 & 25.59 & 27.73 & 23.70 & 28.40 & 28.90 \\
            Baichuan-Omni-1.5     & 25.64 & 32.05 & 28.85 & 23.70 & 23.58 & 29.86 & 24.00 & 24.90 & 29.70 \\
            Qwen-2.5-Omni-7B      & 30.77 & 41.03 & 37.18 & 24.41 & 33.06 & 31.75 & 25.40 & 34.30 & 32.60 \\
            Qwen-3-Omni-30B-A3B   & \textbf{32.05} & \textbf{48.08} & \textbf{48.72} & 23.58 & \textbf{41.23} & \textbf{40.88} & 24.90 & \textbf{42.30} & \textbf{42.10} \\
            \midrule
            Gemini-2.0-Flash      & 25.00 & 48.08 & 45.51 & 22.87 & 48.93 & 44.79 & 23.20 & 48.80 & 44.90 \\
            Gemini-2.5-Flash      & 17.95 & 48.72 & 49.36 & 21.80 & 55.09 & 55.21 & 21.20 & 54.10 & 54.30 \\
            Gemini-2.5-Pro        & \textbf{32.69} & \textbf{57.69} & \textbf{64.10} & \textbf{33.89} & \textbf{66.00} & \textbf{72.16} & \textbf{33.70} & \textbf{64.70} & \textbf{70.90} \\
            \bottomrule
        \end{tabular}
    }
    \label{tab:eval_ablation_audio_part2}
\end{table*}

\subsection{Benchmark Analysis}
In this section, we verify the effectiveness of UNO-Bench on three aspects, the performance of multi-step open-ended question, the performance of dataset compression and the benchmark comparison with other open-source benchmarks.

\subsubsection{Multi-Step Open-Ended Question Analysis}
\label{sec:exp_multi_qa}

In this work, we introduce a new type of evaluation method, multi-step open-ended question, which effectively assess the complex reasoning ability, especially appears in cross-modality understanding.

As shown in Table.\ref{tab:eval_multi_qa}, the experimental results on our multi-step open-ended questions reveal a clear performance stratification among models. Gemini-2.5-Pro establishes itself as the top-tier model with an overall score of 57.32, with Gemini-2.5-Flash (47.08) and Gemini-2.0-Flash (38.56) forming a distinct second tier. Among open-source models, Qwen-3-Omni-30B-A3B emerges as the clear leader with a score of 37.08, significantly outperforming smaller-scale models like Qwen-2.5-Omni-7B (27.72). This starkly illustrates that both advanced architecture and model scale are pivotal factors for success in complex, multi-turn multimodal tasks.

As the depth of questions increases from Q1 to Q3+, most models exhibit a general decline in performance, confirming the effectiveness of our dataset's progressive difficulty. For instance, the leading open-source model, Qwen-3-Omni-30B-A3B, sees its overall score drop from 18.08 on the first question (Q1) to 14.18 (Q2) and further to 11.42 (Q3+). This decay highlights a common challenge for current models in handling long-range dependencies, maintaining conversational context, and performing multi-step reasoning. However, a notable exception is \texttt{Gemini-2.5-Pro}, whose performance on the second question (Q2) surpasses its score on the first (24.48 vs. 23.44), before declining on subsequent questions. This unique pattern suggests a superior ability to utilize the context from the initial turn to enhance its understanding and response in the subsequent turn, a capability not observed in other models.

Reasoning ability remains the key bottleneck that differentiates model performance. For all open-source models and the lower-tier Gemini models, scores on \texttt{Perception} tasks are considerably higher than on \texttt{Reasoning} tasks. The gap is particularly pronounced for Qwen-3-Omni-30B-A3B, which scores 53.8 in Perception but only 32.9 in Reasoning. This indicates that while these models have developed solid foundational perception capabilities, converting this perceptual input into complex logical or causal reasoning remains a major hurdle. Interestingly, Gemini-2.5-Pro is the only model that defies this trend, achieving a higher score in \texttt{Reasoning} (58.1) than in \texttt{Perception} (54.2). This exceptional result demonstrates that state-of-the-art models are beginning to overcome the reasoning bottleneck, showcasing advanced cognitive abilities that are on par with, or even exceed, their perceptual skills. The design of our dataset successfully magnifies this critical capability gap between the SOTA and other models.

\begin{table*}[htbp]
    \caption{Performance on Multi-Step Open-Ended Questions.}
    \centering
    \resizebox{0.9\textwidth}{!}{
        \begin{tabular}{l|cccc|cccc|cccc}
            \toprule
            \multirow{2}{*}{Model} & \multicolumn{4}{c|}{Perception} & \multicolumn{4}{c|}{Reasoning} & \multicolumn{4}{c}{Overall} \\
            \cline{2-13}
            & Q1 & Q2 & Q3+ & All & Q1 & Q2 & Q3+ & All & Q1 & Q2 & Q3+ & All \\
            \midrule
            Baichuan-Omni-1.5         & 15.4  & 8.2  & 5.33  & 25.2  & 9    & 7.25  & 5.75  & 18.9  & 10.28 & 7.44  & 5.7   & 20.16 \\
            MiniCPM-O-2.6             & 20.0  & 6.2  & 11.33 & 29.6  & 9.05 & 9.55  & 8.02  & 22.3  & 11.24 & 8.88  & 8.43  & 23.76 \\
            Qwen-2.5-Omni-3B          & 19.8  & 12.2 & 5.33  & 33.6  & 10.7 & 7.2   & 8.86  & 22.55 & 12.52 & 8.2   & 8.42  & 24.76 \\
            Ming-lite-Omni-1.5        & 19.6  & 12.4 & 4.67  & 33.4  & 10.9 & 8.4   & 7.92  & 23.5  & 12.64 & 9.2   & 7.52  & 25.48 \\
            Qwen-2.5-Omni-7B          & 20.2  & 15.0 & 12.0  & 38.8  & 12.15& 8.99  & 7.83  & 24.95 & 13.76 & 10.2  & 8.35  & 27.72 \\
            Qwen-3-Omni-30B-A3B       & \textbf{25.0} & \textbf{22.8} & \textbf{20.0}  & \textbf{53.8}  & \textbf{16.35} & \textbf{12.01} & \textbf{10.19} & \textbf{32.9}  & \textbf{18.08} & \textbf{14.18} & \textbf{11.42} & \textbf{37.08} \\
            \midrule
            Gemini-2.0-Flash          & 25.2  & 19.4 & 14.67 & 49.0  & 15.5 & 14.05 & 13.02 & 35.95 & 17.44 & 15.12 & 13.22 & 38.56 \\
            Gemini-2.5-Flash          & 31.6  & 22.6 & 12.0  & 57.8  & 18.35& 17.35 & 16.42 & 44.4  & 21.0  & 18.4  & 15.87 & 47.08 \\
            Gemini-2.5-Pro            & \textbf{25.6}  & 22.2 & \textbf{21.33} & \textbf{54.2}  & \textbf{22.9} & \textbf{25.05} & \textbf{19.43} & \textbf{58.1}  & \textbf{23.44} & \textbf{24.48} & \textbf{19.67} & \textbf{57.32} \\
            \bottomrule
        \end{tabular}
    }
    \label{tab:eval_multi_qa}
\end{table*}


\subsubsection{Dataset Compression}
We design a cluster-guided stratified sampling to compress the scale of benchmark. To evaluate the consistency of model ranking and the best size of compression data size, we conduct several experiments to analysis. 

The baseline data set consists of ~8000 samples including 18 open-source benchmarks (e.g. MathVista and MMAU, details see Appendix.\ref{sec:benchmark_for_compression}) and 20 models evaluation results on them, which split into 12/8 on models as training/test set.
Kmeans++\citep{arthur2007k} is used to cluster with K=48. To eliminate the random factor, we conduct 5-fold settings and evaluate 10 times on each setting.

\begin{figure}
    \centering
    \includegraphics[width=0.8\columnwidth]{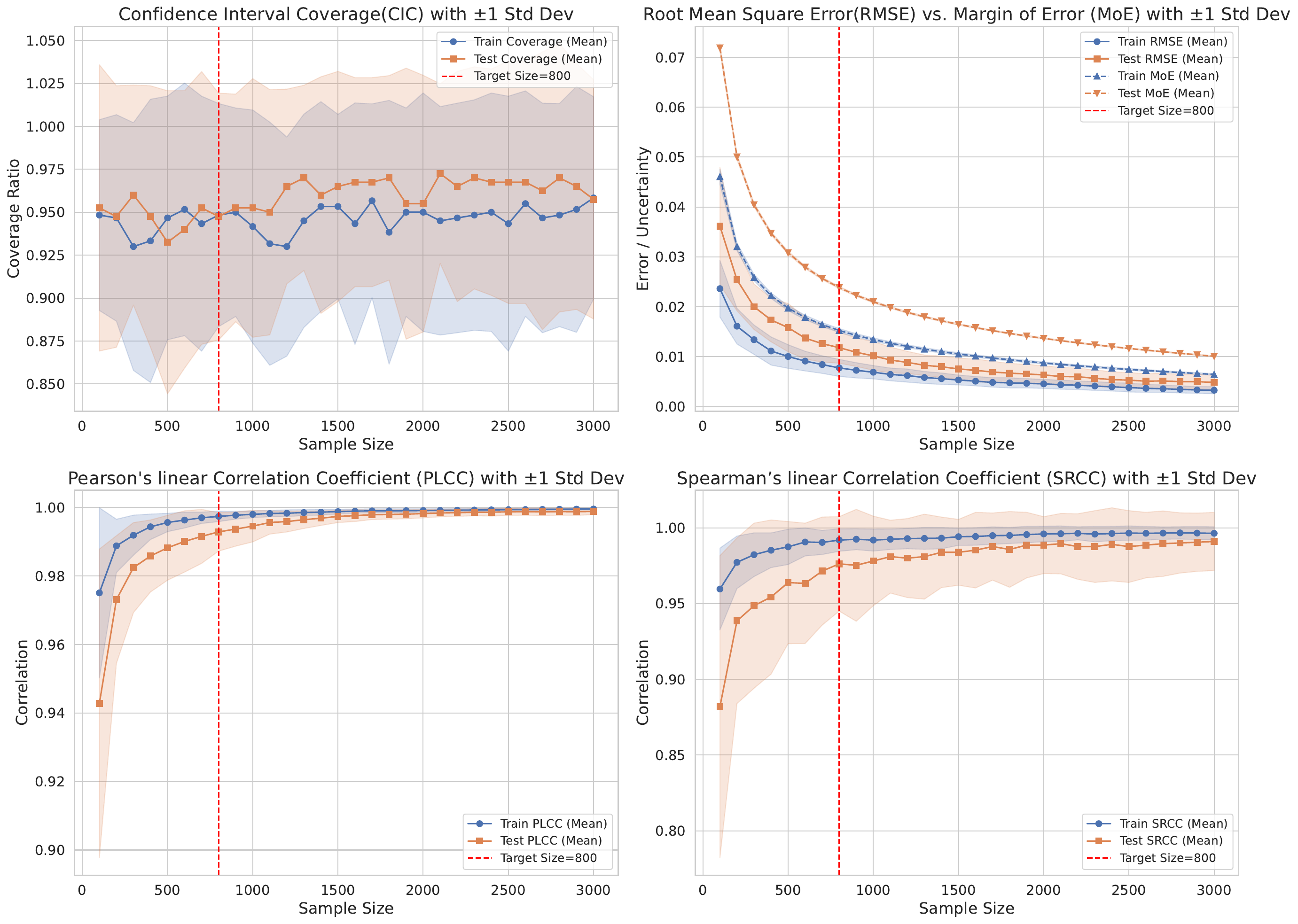}
    \caption{Data compression performance.}
    \label{fig:data_compression}
\end{figure}

The experimental result is shown in Figure.\ref{fig:data_compression}. At a 10\% sampling rate, our method achieved excellent results on test-set. Both \textbf{SRCC} and \textbf{PLCC} exceeded 0.98, indicating near-perfect preservation of ranking and value relationships. The \textbf{RMSE} was below 0.02 with a corresponding \textbf{MoE} of 0.024; together, these values signify high numerical precision and a tight estimation range. Furthermore, the \textbf{CIC} was approximately 95\%, confirming the statistical unbiasedness of the sample.

\subsubsection{Benchmark Comparison}
To ensure the quality of dataset, we conduct quality check on 10\%-20\% random samples in each benchmarks. As shown in Table.\ref{tab:statistic}, UNO-Bench has 100\% accuracy on omni-modal dataset while 98\% questions requires cross-modality to solve. It shows the highest quality among existing omni benchmarks.

\begin{figure}[t]
  \centering
  \begin{minipage}[b]{0.54\textwidth}
    \centering
    \includegraphics[width=\columnwidth]{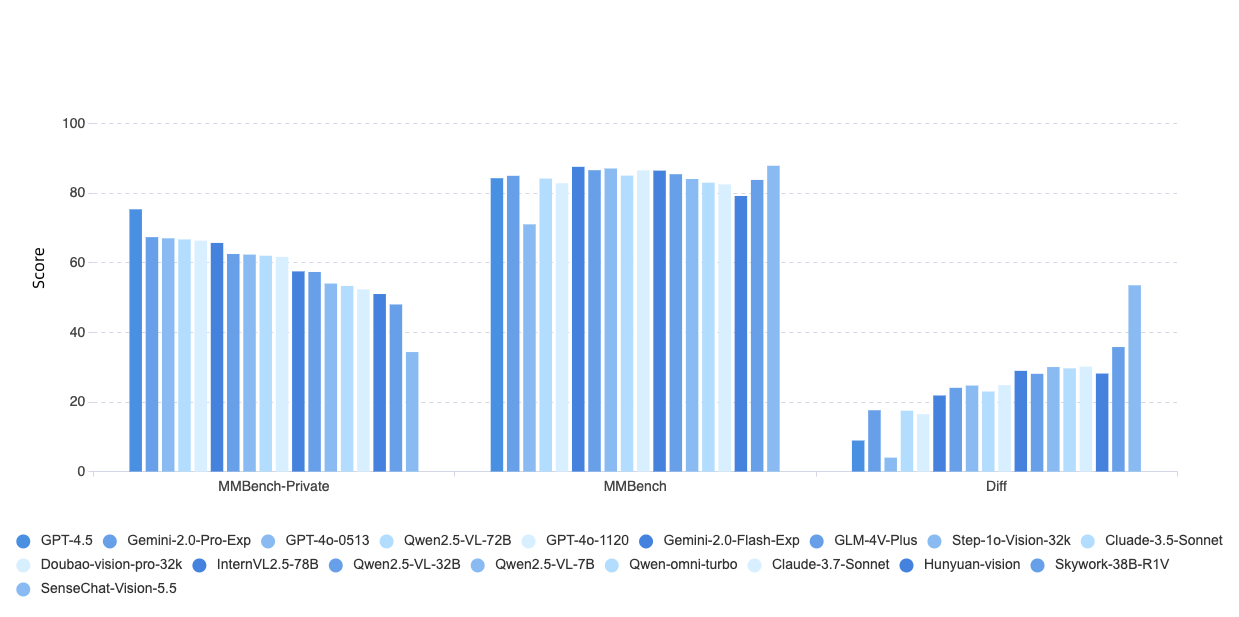}
    \caption{Comparison result of privatization improvement. After improvement, the performances among models are more distinguishable.}
    \label{fig:mmbench_private}
  \end{minipage} 
  \hfill
  \begin{minipage}[b]{0.45\textwidth}
    \centering
    \includegraphics[width=\columnwidth]{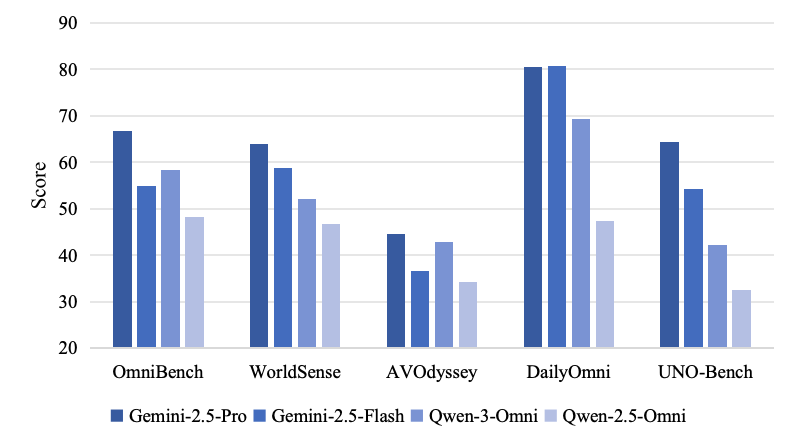}
    \caption{Comparison between omni benchmarks.}
    \label{fig:MMAO-benchmark-comparion}
  \end{minipage}
\end{figure}

An effective benchmark must provide both a clear performance ladder and a meaningful difficulty range, and our UNO-Bench is engineered to deliver on both fronts as shown in Figure.\ref{fig:MMAO-benchmark-comparion}. It excels in discriminability, establishing substantial and remarkably linear intervals of \textasciitilde10-12 points between adjacent models. This superior discriminability comes from a well-calibrated difficulty. UNO-Bench creates a vast 31.9 point performance gap between the top and bottom models, effectively separating their capabilities. This approach avoids the pitfall of being universally difficult, a problem seen in AV-Odyssey where all models are compressed into a narrow, low-scoring band (34-45). By combining a structured performance ladder with a balanced challenge, UNO-Bench serves as a more reliable and insightful tool for gauging genuine progress in the field.

\section{Conclusion}
In this work, we introduce a high quality and diversity benchmark to evaluate omni models comprehensively. 
With unified data framework in UNO-Bench, we found that the omni-modal capability may not simply be a linear superposition of uni-modal capabilities, but rather follows a significant multiplicative relationship. The evaluation results show that it manifests as a bottleneck effect on weak models, while exhibiting synergistic promotion on strong models.
In addition, we found that both uni-modal and omni-modal understanding capability of the Gemini series far surpasses existing open-source omni models. The Gemini-2.5-Pro shows comparable perception capability with human experts but still has a performance gap in reasoning aspect.
Besides better dataset quality and evaluation efficiency, UNO-Bench can provide sufficient metric discriminability and a progressive difficulty scale to drive model capability growth. 

In the future work, we will extend the dataset's scale by the human-in-the-loop automated pipeline and hold a private test set to avoid hacking. The ability coverage also needs to extend to more difficult reasoning tasks like STEM and code. At the same time, the relationship among cross-modals understandings and how to improve them are still exciting problem to explore. Furthermore, our compositional law has been validated on UNO-Bench with uniformly distributed capabilities. Whether this law still holds under different capability distributions remains to be explored.

\section*{Acknowledgement}

We hereby express our appreciation to the LongCat Team EVA Committee for their valuable assistance, guidance, and suggestions throughout the course of this work.

\bibliographystyle{unsrtnat}
\bibliography{custom}

\begin{thebibliography}{56}
\providecommand{\natexlab}[1]{#1}
\providecommand{\url}[1]{\texttt{#1}}
\expandafter\ifx\csname urlstyle\endcsname\relax
  \providecommand{\doi}[1]{doi: #1}\else
  \providecommand{\doi}{doi: \begingroup \urlstyle{rm}\Url}\fi

\bibitem[Li et~al.(2024{\natexlab{a}})Li, Zhang, Ma, Yuan, Zhu, Guo, Liang, Liu, Wang, Yang, Wu, Qu, Shi, Zhang, Yang, Wang, Zhang, Liu, Benetos, Huang, and Lin]{omnibench}
Yizhi Li, Ge~Zhang, Yinghao Ma, Ruibin Yuan, Kang Zhu, Hangyu Guo, Yiming Liang, Jiaheng Liu, Zekun Wang, Jian Yang, Siwei Wu, Xingwei Qu, Jinjie Shi, Xinyue Zhang, Zhenzhu Yang, Xiangzhou Wang, Zhaoxiang Zhang, Zachary Liu, Emmanouil Benetos, Wenhao Huang, and Chenghua Lin.
\newblock Omnibench: Towards the future of universal omni-language models.
\newblock abs/2409.15272, 2024{\natexlab{a}}.

\bibitem[Hong et~al.(2025)Hong, Yan, Cai, Jiang, Hu, and Xie]{worldsense}
Jack Hong, Shilin Yan, Jiayin Cai, Xiaolong Jiang, Yao Hu, and Weidi Xie.
\newblock Worldsense: Evaluating real-world omnimodal understanding for multimodal llms.
\newblock abs/2502.04326, 2025.

\bibitem[Gong et~al.(2024)Gong, Feng, Li, Wang, Cheng, Yang, Han, Wang, Bai, Yang, and Yue]{avodyssey}
Kaixiong Gong, Kaituo Feng, Bohao Li, Yibing Wang, Mofan Cheng, Shijia Yang, Jiaming Han, Benyou Wang, Yutong Bai, Zhuoran Yang, and Xiangyu Yue.
\newblock Av-odyssey bench: Can your multimodal llms really understand audio-visual information?
\newblock abs/2412.02611, 2024.

\bibitem[Liu et~al.(2024{\natexlab{a}})Liu, Duan, Zhang, Li, Zhang, Yuan, Zhao, Wang, He, Liu, Chen, and Lin]{mmbench}
Yuan Liu, Haodong Duan, Yuanhan Zhang, Bo~Li, Songyang Zhang, Yike Yuan, Wangbo Zhao, Jiaqi Wang, Conghui He, Ziwei Liu, Kai Chen, and Dahua Lin.
\newblock Mmbench: is your multi-modal model an all-around player?
\newblock 2024{\natexlab{a}}.

\bibitem[Lu et~al.(2024{\natexlab{a}})Lu, Bansal, Xia, Liu, Li, Hajishirzi, Cheng, Chang, Galley, and Gao]{mathvista}
Pan Lu, Hritik Bansal, Tony Xia, Jiacheng Liu, Chunyuan Li, Hannaneh Hajishirzi, Hao Cheng, Kai-Wei Chang, Michel Galley, and Jianfeng Gao.
\newblock Mathvista: Evaluating mathematical reasoning of foundation models in visual contexts.
\newblock 2024{\natexlab{a}}.

\bibitem[Li et~al.(2024{\natexlab{b}})Li, Wang, He, Li, Wang, Liu, Wang, Xu, Chen, Luo, Wang, and Qiao]{li2024mvbench}
Kunchang Li, Yali Wang, Yinan He, Yizhuo Li, Yi~Wang, Yi~Liu, Zun Wang, Jilan Xu, Guo Chen, Ping Luo, Limin Wang, and Yu~Qiao.
\newblock Mvbench: A comprehensive multi-modal video understanding benchmark, 2024{\natexlab{b}}.
\newblock URL \url{https://arxiv.org/abs/2311.17005}.

\bibitem[Sakshi et~al.(2025)Sakshi, Tyagi, Kumar, Seth, S, Nieto, Duraiswami, Ghosh, and Manocha]{mmau}
Sakshi, Utkarsh Tyagi, Sonal Kumar, Ashish Seth, Ramaneswaran S, Oriol Nieto, Ramani Duraiswami, Sreyan Ghosh, and Dinesh Manocha.
\newblock Mmau: A massive multi-task audio understanding and reasoning benchmark.
\newblock 2025.

\bibitem[Bai et~al.(2025)Bai, Chen, Liu, Wang, Ge, Song, Dang, Wang, Wang, Tang, Zhong, Zhu, Yang, Li, Wan, Wang, Ding, Fu, Xu, Ye, Zhang, Xie, Cheng, Zhang, Yang, Xu, and Lin]{Qwen2.5-VL}
Shuai Bai, Keqin Chen, Xuejing Liu, Jialin Wang, Wenbin Ge, Sibo Song, Kai Dang, Peng Wang, Shijie Wang, Jun Tang, Humen Zhong, Yuanzhi Zhu, Mingkun Yang, Zhaohai Li, Jianqiang Wan, Pengfei Wang, Wei Ding, Zheren Fu, Yiheng Xu, Jiabo Ye, Xi~Zhang, Tianbao Xie, Zesen Cheng, Hang Zhang, Zhibo Yang, Haiyang Xu, and Junyang Lin.
\newblock Qwen2.5-vl technical report.
\newblock \emph{arXiv preprint arXiv:2502.13923}, 2025.

\bibitem[Xiaomi(2025)]{coreteam2025mimovltechnicalreport}
LLM-Core-Team Xiaomi.
\newblock Mimo-vl technical report, 2025.
\newblock URL \url{https://arxiv.org/abs/2506.03569}.

\bibitem[Zeng et~al.(2025)Zeng, Lv, Zheng, Hou, Chen, Xie, Wang, Yin, Zeng, Zhang, et~al.]{zeng2025glm}
Aohan Zeng, Xin Lv, Qinkai Zheng, Zhenyu Hou, Bin Chen, Chengxing Xie, Cunxiang Wang, Da~Yin, Hao Zeng, Jiajie Zhang, et~al.
\newblock {GLM-4.5}: Agentic, reasoning, and coding ({ARC}) foundation models.
\newblock \emph{arXiv preprint arXiv:2508.06471}, 2025.

\bibitem[Ding et~al.(2025)Ding, Ju, Leng, Liu, Liu, Shang, Shen, Song, Tan, Tang, et~al.]{ding2025kimi}
Ding Ding, Zeqian Ju, Yichong Leng, Songxiang Liu, Tong Liu, Zeyu Shang, Kai Shen, Wei Song, Xu~Tan, Heyi Tang, et~al.
\newblock Kimi-audio technical report.
\newblock \emph{arXiv preprint arXiv:2504.18425}, 2025.

\bibitem[Wu et~al.(2025)Wu, Yan, Hu, Yi, Feng, Tian, Shen, Yu, Zhang, Li, et~al.]{wu2025step}
Boyong Wu, Chao Yan, Chen Hu, Cheng Yi, Chengli Feng, Fei Tian, Feiyu Shen, Gang Yu, Haoyang Zhang, Jingbei Li, et~al.
\newblock Step-audio 2 technical report.
\newblock \emph{arXiv preprint arXiv:2507.16632}, 2025.

\bibitem[Wang et~al.(2024{\natexlab{a}})Wang, Pan, Shi, Lu, Ren, Zhou, Zhan, and Li]{wang2024mathvision}
Ke~Wang, Junting Pan, Weikang Shi, Zimu Lu, Houxing Ren, Aojun Zhou, Mingjie Zhan, and Hongsheng Li.
\newblock Measuring multimodal mathematical reasoning with math-vision dataset.
\newblock In \emph{The Thirty-eight Conference on Neural Information Processing Systems Datasets and Benchmarks Track}, 2024{\natexlab{a}}.
\newblock URL \url{https://openreview.net/forum?id=QWTCcxMpPA}.

\bibitem[Wang et~al.(2024{\natexlab{b}})Wang, Xia, He, Chen, Liu, Zhu, Liang, Wu, Liu, Malladi, Chevalier, Arora, and Chen]{wang2024charxiv}
Zirui Wang, Mengzhou Xia, Luxi He, Howard Chen, Yitao Liu, Richard Zhu, Kaiqu Liang, Xindi Wu, Haotian Liu, Sadhika Malladi, Alexis Chevalier, Sanjeev Arora, and Danqi Chen.
\newblock Charxiv: Charting gaps in realistic chart understanding in multimodal llms.
\newblock \emph{arXiv preprint arXiv:2406.18521}, 2024{\natexlab{b}}.

\bibitem[Liu et~al.(2024{\natexlab{b}})Liu, Li, Huang, Yang, Yu, Li, Yin, Liu, Jin, and Bai]{liu2024ocrbench}
Yuliang Liu, Zhang Li, Mingxin Huang, Biao Yang, Wenwen Yu, Chunyuan Li, Xu-Cheng Yin, Cheng-Lin Liu, Lianwen Jin, and Xiang Bai.
\newblock Ocrbench: on the hidden mystery of ocr in large multimodal models.
\newblock \emph{Science China Information Sciences}, 67\penalty0 (12), December 2024{\natexlab{b}}.
\newblock ISSN 1869-1919.
\newblock \doi{10.1007/s11432-024-4235-6}.
\newblock URL \url{http://dx.doi.org/10.1007/s11432-024-4235-6}.

\bibitem[Mathew et~al.(2021)Mathew, Karatzas, and Jawahar]{mathew2021docvqa}
Minesh Mathew, Dimosthenis Karatzas, and C.~V. Jawahar.
\newblock Docvqa: A dataset for vqa on document images, 2021.
\newblock URL \url{https://arxiv.org/abs/2007.00398}.

\bibitem[Ouyang et~al.(2024)Ouyang, Qu, Zhou, Zhu, Zhang, Lin, Wang, Zhao, Jiang, Zhao, Shi, Wu, Chu, Liu, Li, Xu, Zhang, Shi, Tu, and He]{ouyang2024omnidocbench}
Linke Ouyang, Yuan Qu, Hongbin Zhou, Jiawei Zhu, Rui Zhang, Qunshu Lin, Bin Wang, Zhiyuan Zhao, Man Jiang, Xiaomeng Zhao, Jin Shi, Fan Wu, Pei Chu, Minghao Liu, Zhenxiang Li, Chao Xu, Bo~Zhang, Botian Shi, Zhongying Tu, and Conghui He.
\newblock Omnidocbench: Benchmarking diverse pdf document parsing with comprehensive annotations, 2024.
\newblock URL \url{https://arxiv.org/abs/2412.07626}.

\bibitem[Wu et~al.(2024)Wu, Li, Chen, and Li]{wu2024longvideobench}
Haoning Wu, Dongxu Li, Bei Chen, and Junnan Li.
\newblock Longvideobench: A benchmark for long-context interleaved video-language understanding, 2024.
\newblock URL \url{https://arxiv.org/abs/2407.15754}.

\bibitem[Liu et~al.(2024{\natexlab{c}})Liu, Li, Liu, Wang, Ren, Li, Chen, Sun, and Hou]{liu2024tempcompass}
Yuanxin Liu, Shicheng Li, Yi~Liu, Yuxiang Wang, Shuhuai Ren, Lei Li, Sishuo Chen, Xu~Sun, and Lu~Hou.
\newblock Tempcompass: Do video llms really understand videos?
\newblock \emph{arXiv preprint arXiv: 2403.00476}, 2024{\natexlab{c}}.

\bibitem[xAI(2023)]{realworldqa}
xAI.
\newblock Realworldqa.
\newblock \url{https://huggingface.co/datasets/xai-org/RealworldQA}, 2023.
\newblock Version 1.0, Accessed: 2024.

\bibitem[Xiao et~al.(2021)Xiao, Shang, Yao, and Chua]{xiao2021nextqa}
Junbin Xiao, Xindi Shang, Angela Yao, and Tat-Seng Chua.
\newblock Next-qa: Next phase of question-answering to explaining temporal actions.
\newblock In \emph{Proceedings of the IEEE/CVF Conference on Computer Vision and Pattern Recognition (CVPR)}, pages 9777--9786, June 2021.

\bibitem[Huang et~al.(2025)Huang, Li, Li, Wang, Zeng, Liang, Wu, Chen, Li, and Wang]{huang2025ovbench}
Zhenpeng Huang, Xinhao Li, Jiaqi Li, Jing Wang, Xiangyu Zeng, Cheng Liang, Tao Wu, Xi~Chen, Liang Li, and Limin Wang.
\newblock Online video understanding: Ovbench and videochat-online, 2025.
\newblock URL \url{https://arxiv.org/abs/2501.00584}.

\bibitem[Hu et~al.(2025)Hu, Wu, Pu, Xiao, Zhang, Yue, Li, and Liu]{hu2025videommmu}
Kairui Hu, Penghao Wu, Fanyi Pu, Wang Xiao, Yuanhan Zhang, Xiang Yue, Bo~Li, and Ziwei Liu.
\newblock Video-mmmu: Evaluating knowledge acquisition from multi-discipline professional videos.
\newblock 2025.
\newblock URL \url{https://arxiv.org/abs/2501.13826}.

\bibitem[Fu et~al.(2024)Fu, Dai, Luo, Li, Ren, Zhang, Wang, Zhou, Shen, Zhang, et~al.]{fu2024videomme}
Chaoyou Fu, Yuhan Dai, Yondong Luo, Lei Li, Shuhuai Ren, Renrui Zhang, Zihan Wang, Chenyu Zhou, Yunhang Shen, Mengdan Zhang, et~al.
\newblock Video-mme: The first-ever comprehensive evaluation benchmark of multi-modal llms in video analysis.
\newblock \emph{arXiv preprint arXiv:2405.21075}, 2024.

\bibitem[Ardila et~al.(2019)Ardila, Branson, Davis, Henretty, Kohler, Meyer, Morais, Saunders, Tyers, and Weber]{ardila2019common}
Rosana Ardila, Megan Branson, Kelly Davis, Michael Henretty, Michael Kohler, Josh Meyer, Reuben Morais, Lindsay Saunders, Francis~M Tyers, and Gregor Weber.
\newblock Common voice: A massively-multilingual speech corpus.
\newblock \emph{arXiv preprint arXiv:1912.06670}, 2019.

\bibitem[Wang et~al.(2021)Wang, Wu, Gu, and Pino]{wang2021covost}
Changhan Wang, Anne Wu, Jiatao Gu, and Juan Pino.
\newblock Covost 2 and massively multilingual speech translation.
\newblock In \emph{Interspeech}, volume 2021, pages 2247--2251, 2021.

\bibitem[Yang et~al.(2024)Yang, Xu, Liu, Chu, Jiang, Zhou, Leng, Lv, Zhao, Zhou, et~al.]{yang2024air}
Qian Yang, Jin Xu, Wenrui Liu, Yunfei Chu, Ziyue Jiang, Xiaohuan Zhou, Yichong Leng, Yuanjun Lv, Zhou Zhao, Chang Zhou, et~al.
\newblock Air-bench: Benchmarking large audio-language models via generative comprehension.
\newblock \emph{arXiv preprint arXiv:2402.07729}, 2024.

\bibitem[Ao et~al.(2024)Ao, Wang, Tian, Chen, Zhang, Lu, Wang, Li, and Wu]{ao2024sd}
Junyi Ao, Yuancheng Wang, Xiaohai Tian, Dekun Chen, Jun Zhang, Lu~Lu, Yuxuan Wang, Haizhou Li, and Zhizheng Wu.
\newblock Sd-eval: A benchmark dataset for spoken dialogue understanding beyond words.
\newblock \emph{Advances in Neural Information Processing Systems}, 37:\penalty0 56898--56918, 2024.

\bibitem[Comanici et~al.(2025)Comanici, Bieber, Schaekermann, Pasupat, Sachdeva, Dhillon, Blistein, Ram, Zhang, Rosen, et~al.]{comanici2025gemini}
Gheorghe Comanici, Eric Bieber, Mike Schaekermann, Ice Pasupat, Noveen Sachdeva, Inderjit Dhillon, Marcel Blistein, Ori Ram, Dan Zhang, Evan Rosen, et~al.
\newblock Gemini 2.5: Pushing the frontier with advanced reasoning, multimodality, long context, and next generation agentic capabilities.
\newblock \emph{arXiv preprint arXiv:2507.06261}, 2025.

\bibitem[Xu et~al.(2025{\natexlab{a}})Xu, Guo, Hu, Chu, Wang, He, Wang, Shi, He, Zhu, Lv, Wang, Guo, Wang, Ma, Zhang, Zhang, Hao, Guo, Yang, Zhang, Ma, Wei, Bai, Chen, Liu, Wang, Yang, Liu, Ren, Zheng, Men, Zhou, Yu, Yang, Yu, Zhou, and Lin]{qwen3omni}
Jin Xu, Zhifang Guo, Hangrui Hu, Yunfei Chu, Xiong Wang, Jinzheng He, Yuxuan Wang, Xian Shi, Ting He, Xinfa Zhu, Yuanjun Lv, Yongqi Wang, Dake Guo, He~Wang, Linhan Ma, Pei Zhang, Xinyu Zhang, Hongkun Hao, Zishan Guo, Baosong Yang, Bin Zhang, Ziyang Ma, Xipin Wei, Shuai Bai, Keqin Chen, Xuejing Liu, Peng Wang, Mingkun Yang, Dayiheng Liu, Xingzhang Ren, Bo~Zheng, Rui Men, Fan Zhou, Bowen Yu, Jianxin Yang, Le~Yu, Jingren Zhou, and Junyang Lin.
\newblock Qwen3-omni technical report.
\newblock 2025{\natexlab{a}}.

\bibitem[AI et~al.(2025)AI, Gong, Zou, Zheng, Zhou, Yan, Jin, Shen, Zheng, Wang, Xu, Yao, Zhou, Chen, Sun, Liu, Zhu, Peng, Ji, Song, Ren, Wang, Ru, Xie, Tan, Xue, Wang, Bai, Gao, Chen, Guo, Zhang, Xu, Liu, Xiong, Gao, Liu, Li, Chai, Xiao, Wang, Chen, Lu, Li, Dong, Yu, Yuan, Gao, Sun, Chen, Wu, Lyu, Ma, Feng, Fang, Qiu, Huang, and He]{mingomni}
Inclusion AI, Biao Gong, Cheng Zou, Chuanyang Zheng, Chunluan Zhou, Canxiang Yan, Chunxiang Jin, Chunjie Shen, Dandan Zheng, Fudong Wang, Furong Xu, GuangMing Yao, Jun Zhou, Jingdong Chen, Jianxin Sun, Jiajia Liu, Jianjiang Zhu, Jun Peng, Kaixiang Ji, Kaiyou Song, Kaimeng Ren, Libin Wang, Lixiang Ru, Lele Xie, Longhua Tan, Lyuxin Xue, Lan Wang, Mochen Bai, Ning Gao, Pei Chen, Qingpei Guo, Qinglong Zhang, Qiang Xu, Rui Liu, Ruijie Xiong, Sirui Gao, Tinghao Liu, Taisong Li, Weilong Chai, Xinyu Xiao, Xiaomei Wang, Xiaoxue Chen, Xiao Lu, Xiaoyu Li, Xingning Dong, Xuzheng Yu, Yi~Yuan, Yuting Gao, Yunxiao Sun, Yipeng Chen, Yifei Wu, Yongjie Lyu, Ziping Ma, Zipeng Feng, Zhijiang Fang, Zhihao Qiu, Ziyuan Huang, and Zhengyu He.
\newblock Ming-omni: A unified multimodal model for perception and generation.
\newblock 2025.

\bibitem[Li et~al.(2025)Li, Liu, Zhang, Zhang, Chen, Li, Li, Liu, Ming, Dong, Pan, Li, Fang, Kuang, Wang, Zhu, Zhang, Guo, Zhang, Wang, Ding, Song, Li, Huo, Liang, Zhang, Wu, Zhao, Xiong, Wu, Ye, Lu, Li, Zhang, Zhou, Chen, Su, Zhang, Chen, Dong, Nie, Wu, Xiao, Li, Dang, Zhang, Sun, Wu, Yang, Lin, Ma, Wu, li, Yang, Liu, Zhang, Chen, Ai, Zhang, Chen, Huang, Li, Luo, Duan, Zhu, Xiao, Su, Pu, Wang, Jia, Zhang, Ai, Wang, Qiao, Zhang, Shen, Yang, Zhen, Zhou, Chen, Li, Zhu, Lu, Zhao, Liang, Li, Qin, Sun, Xu, Sun, Lin, Zhou, and Chen]{li2025baichuanomni15technicalreport}
Yadong Li, Jun Liu, Tao Zhang, Tao Zhang, Song Chen, Tianpeng Li, Zehuan Li, Lijun Liu, Lingfeng Ming, Guosheng Dong, Da~Pan, Chong Li, Yuanbo Fang, Dongdong Kuang, Mingrui Wang, Chenglin Zhu, Youwei Zhang, Hongyu Guo, Fengyu Zhang, Yuran Wang, Bowen Ding, Wei Song, Xu~Li, Yuqi Huo, Zheng Liang, Shusen Zhang, Xin Wu, Shuai Zhao, Linchu Xiong, Yozhen Wu, Jiahui Ye, Wenhao Lu, Bowen Li, Yan Zhang, Yaqi Zhou, Xin Chen, Lei Su, Hongda Zhang, Fuzhong Chen, Xuezhen Dong, Na~Nie, Zhiying Wu, Bin Xiao, Ting Li, Shunya Dang, Ping Zhang, Yijia Sun, Jincheng Wu, Jinjie Yang, Xionghai Lin, Zhi Ma, Kegeng Wu, Jia li, Aiyuan Yang, Hui Liu, Jianqiang Zhang, Xiaoxi Chen, Guangwei Ai, Wentao Zhang, Yicong Chen, Xiaoqin Huang, Kun Li, Wenjing Luo, Yifei Duan, Lingling Zhu, Ran Xiao, Zhe Su, Jiani Pu, Dian Wang, Xu~Jia, Tianyu Zhang, Mengyu Ai, Mang Wang, Yujing Qiao, Lei Zhang, Yanjun Shen, Fan Yang, Miao Zhen, Yijie Zhou, Mingyang Chen, Fei Li, Chenzheng Zhu, Keer Lu, Yaqi Zhao, Hao Liang, Youquan Li, Yanzhao Qin, Linzhuang
  Sun, Jianhua Xu, Haoze Sun, Mingan Lin, Zenan Zhou, and Weipeng Chen.
\newblock Baichuan-omni-1.5 technical report, 2025.
\newblock URL \url{https://arxiv.org/abs/2501.15368}.

\bibitem[Zhou et~al.(2025)Zhou, Wang, and Wu]{zhou2025dailyomni}
Ziwei Zhou, Rui Wang, and Zuxuan Wu.
\newblock Daily-omni: Towards audio-visual reasoning with temporal alignment across modalities, 2025.
\newblock URL \url{https://arxiv.org/abs/2505.17862}.

\bibitem[Zhao et~al.(2025)Zhao, Xie, Zhang, Gan, Long, Hu, Hu, Chen, Li, Song, Xu, Wang, Pan, Shangguan, Tang, Liang, Liu, Zhao, and Cohan]{zhao2025mmvu}
Yilun Zhao, Lujing Xie, Haowei Zhang, Guo Gan, Yitao Long, Zhiyuan Hu, Tongyan Hu, Weiyuan Chen, Chuhan Li, Junyang Song, Zhijian Xu, Chengye Wang, Weifeng Pan, Ziyao Shangguan, Xiangru Tang, Zhenwen Liang, Yixin Liu, Chen Zhao, and Arman Cohan.
\newblock Mmvu: Measuring expert-level multi-discipline video understanding, 2025.
\newblock URL \url{https://arxiv.org/abs/2501.12380}.

\bibitem[Chen et~al.(2025)Chen, Li, Shi, Hu, Luo, Wang, and Zhang]{chen2025videovista}
Xinyu Chen, Yunxin Li, Haoyuan Shi, Baotian Hu, Wenhan Luo, Yaowei Wang, and Min Zhang.
\newblock Videovista-culturallingo: 360 horizons-bridging cultures, languages, and domains in video comprehension.
\newblock \emph{arXiv preprint arXiv:2504.17821}, 2025.

\bibitem[Xu et~al.(2024)Xu, Wang, Fan, and Liu]{xu2024leakage}
Ruijie Xu, Zengzhi Wang, Run-Ze Fan, and Pengfei Liu.
\newblock Benchmarking benchmark leakage in large language models, 2024.
\newblock URL \url{https://arxiv.org/abs/2404.18824}.

\bibitem[Arthur and Vassilvitskii(2007)]{arthur2007k}
David Arthur and Sergei Vassilvitskii.
\newblock k-means++: The advantages of careful seeding.
\newblock In \emph{Proceedings of the Eighteenth Annual ACM-SIAM Symposium on Discrete Algorithms}, pages 1027--1035, 2007.

\bibitem[Lightman et~al.(2023)Lightman, Kosaraju, Burda, Edwards, Baker, Lee, Leike, Schulman, Sutskever, and Cobbe]{lightman2023let}
Hunter Lightman, Vineet Kosaraju, Yuri Burda, Harrison Edwards, Bowen Baker, Teddy Lee, Jan Leike, John Schulman, Ilya Sutskever, and Karl Cobbe.
\newblock Let's verify step by step.
\newblock In \emph{The Twelfth International Conference on Learning Representations}, 2023.

\bibitem[Reddy et~al.(2019)Reddy, Chen, and Manning]{reddy2019coqa}
Siva Reddy, Danqi Chen, and Christopher~D Manning.
\newblock Coqa: A conversational question answering challenge.
\newblock \emph{Transactions of the Association for Computational Linguistics}, 7:\penalty0 249--266, 2019.

\bibitem[Yang et~al.(2025)Yang, Li, Yang, Zhang, Hui, Zheng, Yu, Gao, Huang, Lv, Zheng, Liu, Zhou, Huang, Hu, Ge, Wei, Lin, Tang, Yang, Tu, Zhang, Yang, Yang, Zhou, Zhou, Lin, Dang, Bao, Yang, Yu, Deng, Li, Xue, Li, Zhang, Wang, Zhu, Men, Gao, Liu, Luo, Li, Tang, Yin, Ren, Wang, Zhang, Ren, Fan, Su, Zhang, Zhang, Wan, Liu, Wang, Cui, Zhang, Zhou, and Qiu]{qwen3}
An~Yang, Anfeng Li, Baosong Yang, Beichen Zhang, Binyuan Hui, Bo~Zheng, Bowen Yu, Chang Gao, Chengen Huang, Chenxu Lv, Chujie Zheng, Dayiheng Liu, Fan Zhou, Fei Huang, Feng Hu, Hao Ge, Haoran Wei, Huan Lin, Jialong Tang, Jian Yang, Jianhong Tu, Jianwei Zhang, Jianxin Yang, Jiaxi Yang, Jing Zhou, Jingren Zhou, Junyang Lin, Kai Dang, Keqin Bao, Kexin Yang, Le~Yu, Lianghao Deng, Mei Li, Mingfeng Xue, Mingze Li, Pei Zhang, Peng Wang, Qin Zhu, Rui Men, Ruize Gao, Shixuan Liu, Shuang Luo, Tianhao Li, Tianyi Tang, Wenbiao Yin, Xingzhang Ren, Xinyu Wang, Xinyu Zhang, Xuancheng Ren, Yang Fan, Yang Su, Yichang Zhang, Yinger Zhang, Yu~Wan, Yuqiong Liu, Zekun Wang, Zeyu Cui, Zhenru Zhang, Zhipeng Zhou, and Zihan Qiu.
\newblock Qwen3 technical report.
\newblock \emph{arXiv preprint arXiv:2505.09388}, 2025.

\bibitem[Xu et~al.(2025{\natexlab{b}})Xu, Guo, He, Hu, He, Bai, Chen, Wang, Fan, Dang, Zhang, Wang, Chu, and Lin]{xu2025qwen25omnitechnicalreport}
Jin Xu, Zhifang Guo, Jinzheng He, Hangrui Hu, Ting He, Shuai Bai, Keqin Chen, Jialin Wang, Yang Fan, Kai Dang, Bin Zhang, Xiong Wang, Yunfei Chu, and Junyang Lin.
\newblock Qwen2.5-omni technical report, 2025{\natexlab{b}}.
\newblock URL \url{https://arxiv.org/abs/2503.20215}.

\bibitem[Yao et~al.(2024)Yao, Yu, Zhang, Wang, Cui, Zhu, Cai, Li, Zhao, He, et~al.]{yao2024minicpm}
Yuan Yao, Tianyu Yu, Ao~Zhang, Chongyi Wang, Junbo Cui, Hongji Zhu, Tianchi Cai, Haoyu Li, Weilin Zhao, Zhihui He, et~al.
\newblock Minicpm-v: A gpt-4v level mllm on your phone.
\newblock \emph{arXiv preprint arXiv:2408.01800}, 2024.

\bibitem[Chaoyou et~al.(2023)Chaoyou, Peixian, Yunhang, Yulei, Mengdan, Xu, Jinrui, Xiawu, Ke, Xing, et~al.]{chaoyou2023mme}
Fu~Chaoyou, Chen Peixian, Shen Yunhang, Qin Yulei, Zhang Mengdan, Lin Xu, Yang Jinrui, Zheng Xiawu, Li~Ke, Sun Xing, et~al.
\newblock Mme: A comprehensive evaluation benchmark for multimodal large language models.
\newblock \emph{arXiv preprint arXiv:2306.13394}, 3, 2023.

\bibitem[Li et~al.(2023)Li, Wang, Wang, Ge, Ge, and Shan]{li2023seed}
Bohao Li, Rui Wang, Guangzhi Wang, Yuying Ge, Yixiao Ge, and Ying Shan.
\newblock Seed-bench: Benchmarking multimodal llms with generative comprehension.
\newblock \emph{arXiv preprint arXiv:2307.16125}, 2023.

\bibitem[Liu et~al.(2024{\natexlab{d}})Liu, Wei, Chen, Kong, Ge, Zhu, Zhao, Sun, Han, and Zhang]{liu2024focus}
Chenglong Liu, Haoran Wei, Jinyue Chen, Lingyu Kong, Zheng Ge, Zining Zhu, Liang Zhao, Jianjian Sun, Chunrui Han, and Xiangyu Zhang.
\newblock Focus anywhere for fine-grained multi-page document understanding.
\newblock \emph{arXiv preprint arXiv:2405.14295}, 2024{\natexlab{d}}.

\bibitem[Ye et~al.(2023)Ye, Hu, Xu, Ye, Yan, Dan, Zhao, Xu, Li, Tian, et~al.]{ye2023mplug}
Jiabo Ye, Anwen Hu, Haiyang Xu, Qinghao Ye, Ming Yan, Yuhao Dan, Chenlin Zhao, Guohai Xu, Chenliang Li, Junfeng Tian, et~al.
\newblock mplug-docowl: Modularized multimodal large language model for document understanding.
\newblock \emph{arXiv preprint arXiv:2307.02499}, 2023.

\bibitem[Yue et~al.(2024)Yue, Ni, Zhang, Zheng, Liu, Zhang, Stevens, Jiang, Ren, Sun, Wei, Yu, Yuan, Sun, Yin, Zheng, Yang, Liu, Huang, Sun, Su, and Chen]{yue2023mmmu}
Xiang Yue, Yuansheng Ni, Kai Zhang, Tianyu Zheng, Ruoqi Liu, Ge~Zhang, Samuel Stevens, Dongfu Jiang, Weiming Ren, Yuxuan Sun, Cong Wei, Botao Yu, Ruibin Yuan, Renliang Sun, Ming Yin, Boyuan Zheng, Zhenzhu Yang, Yibo Liu, Wenhao Huang, Huan Sun, Yu~Su, and Wenhu Chen.
\newblock Mmmu: A massive multi-discipline multimodal understanding and reasoning benchmark for expert agi.
\newblock In \emph{Proceedings of CVPR}, 2024.

\bibitem[Yue et~al.(2025)Yue, Zheng, Ni, Wang, Zhang, Tong, Sun, Yu, Zhang, Sun, Su, Chen, and Neubig]{yue2025mmmupro}
Xiang Yue, Tianyu Zheng, Yuansheng Ni, Yubo Wang, Kai Zhang, Shengbang Tong, Yuxuan Sun, Botao Yu, Ge~Zhang, Huan Sun, Yu~Su, Wenhu Chen, and Graham Neubig.
\newblock Mmmu-pro: A more robust multi-discipline multimodal understanding benchmark, 2025.
\newblock URL \url{https://arxiv.org/abs/2409.02813}.

\bibitem[Ge et~al.(2024)Ge, Xinrun, Bei, Yiming, Tongxu, Tianyu, Kang, Yuyang, Chunpu, Shuyue, Haoran, Xingwei, Junjie, Ruibin, Yizhi, Zekun, Yudong, Yu-Hsuan, Fengji, Chenghua, Wenhao, and Jie]{zhang2024cmmmu}
Zhang Ge, Du~Xinrun, Chen Bei, Liang Yiming, Luo Tongxu, Zheng Tianyu, Zhu Kang, Cheng Yuyang, Xu~Chunpu, Guo Shuyue, Zhang Haoran, Qu~Xingwei, Wang Junjie, Yuan Ruibin, Li~Yizhi, Wang Zekun, Liu Yudong, Tsai Yu-Hsuan, Zhang Fengji, Lin Chenghua, Huang Wenhao, and Fu~Jie.
\newblock Cmmmu: A chinese massive multi-discipline multimodal understanding benchmark.
\newblock \emph{arXiv preprint arXiv:2401.20847}, 2024.

\bibitem[Lu et~al.(2024{\natexlab{b}})Lu, Bansal, Xia, Liu, Li, Hajishirzi, Cheng, Chang, Galley, and Gao]{lu2024mathvista}
Pan Lu, Hritik Bansal, Tony Xia, Jiacheng Liu, Chunyuan Li, Hannaneh Hajishirzi, Hao Cheng, Kai-Wei Chang, Michel Galley, and Jianfeng Gao.
\newblock Mathvista: Evaluating mathematical reasoning of foundation models in visual contexts.
\newblock In \emph{International Conference on Learning Representations (ICLR)}, 2024{\natexlab{b}}.

\bibitem[Team et~al.(2025)Team, Du, Gao, Xing, Jiang, Chen, Li, Xiao, Du, Liao, et~al.]{team2025kimi}
Kimi Team, Angang Du, Bofei Gao, Bowei Xing, Changjiu Jiang, Cheng Chen, Cheng Li, Chenjun Xiao, Chenzhuang Du, Chonghua Liao, et~al.
\newblock Kimi k1. 5: Scaling reinforcement learning with llms.
\newblock \emph{arXiv preprint arXiv:2501.12599}, 2025.

\bibitem[Ye et~al.(2024)Ye, Wang, Li, Deng, Li, Li, Duan, Huang, Su, Wang, et~al.]{ye2024gmai}
Jin Ye, Guoan Wang, Yanjun Li, Zhongying Deng, Wei Li, Tianbin Li, Haodong Duan, Ziyan Huang, Yanzhou Su, Benyou Wang, et~al.
\newblock Gmai-mmbench: A comprehensive multimodal evaluation benchmark towards general medical ai.
\newblock \emph{Advances in Neural Information Processing Systems}, 37:\penalty0 94327--94427, 2024.

\bibitem[Kazemi et~al.(2024)Kazemi, Dikkala, Anand, Devic, Dasgupta, Liu, Fatemi, Awasthi, Gollapudi, Guo, et~al.]{kazemi2024remi}
Mehran Kazemi, Nishanth Dikkala, Ankit Anand, Petar Devic, Ishita Dasgupta, Fangyu Liu, Bahare Fatemi, Pranjal Awasthi, Sreenivas Gollapudi, Dee Guo, et~al.
\newblock Remi: A dataset for reasoning with multiple images.
\newblock \emph{Advances in Neural Information Processing Systems}, 37:\penalty0 60088--60109, 2024.

\bibitem[Wang et~al.(2024{\natexlab{c}})Wang, Fu, Huang, Li, Liu, Liu, Ma, Xu, Zhou, Zhang, et~al.]{wang2024muirbench}
Fei Wang, Xingyu Fu, James~Y Huang, Zekun Li, Qin Liu, Xiaogeng Liu, Mingyu~Derek Ma, Nan Xu, Wenxuan Zhou, Kai Zhang, et~al.
\newblock Muirbench: A comprehensive benchmark for robust multi-image understanding.
\newblock \emph{arXiv preprint arXiv:2406.09411}, 2024{\natexlab{c}}.

\bibitem[Wang et~al.(2025)Wang, Wu, Li, Yang, Chen, Zhang, and Meng]{wang2025mmsu}
Dingdong Wang, Jincenzi Wu, Junan Li, Dongchao Yang, Xueyuan Chen, Tianhua Zhang, and Helen Meng.
\newblock Mmsu: A massive multi-task spoken language understanding and reasoning benchmark.
\newblock \emph{arXiv preprint arXiv:2506.04779}, 2025.

\bibitem[Faisal et~al.(2021)Faisal, Keshava, ibn Alam, and Anastasopoulos]{faisal-etal-21-sdqa}
Fahim Faisal, Sharlina Keshava, Md~Mahfuz ibn Alam, and Antonios Anastasopoulos.
\newblock {SD-QA}: {S}poken {D}ialectal {Q}uestion {A}nswering for the {R}eal {W}orld.
\newblock In \emph{Findings of the 2021 Conference on Empirical Methods in Natural Language Processing (EMNLP Findings)}. Association for Computational Linguistics, November 2021.
\newblock URL \url{https://arxiv.org/abs/2109.12072}.

\end{thebibliography}

\clearpage

\appendix

\section{Benchmarks Utilized in Dataset Compression} 
\label{sec:benchmark_for_compression} 
\begin{figure}
    \centering
    \includegraphics[width=0.6\linewidth]{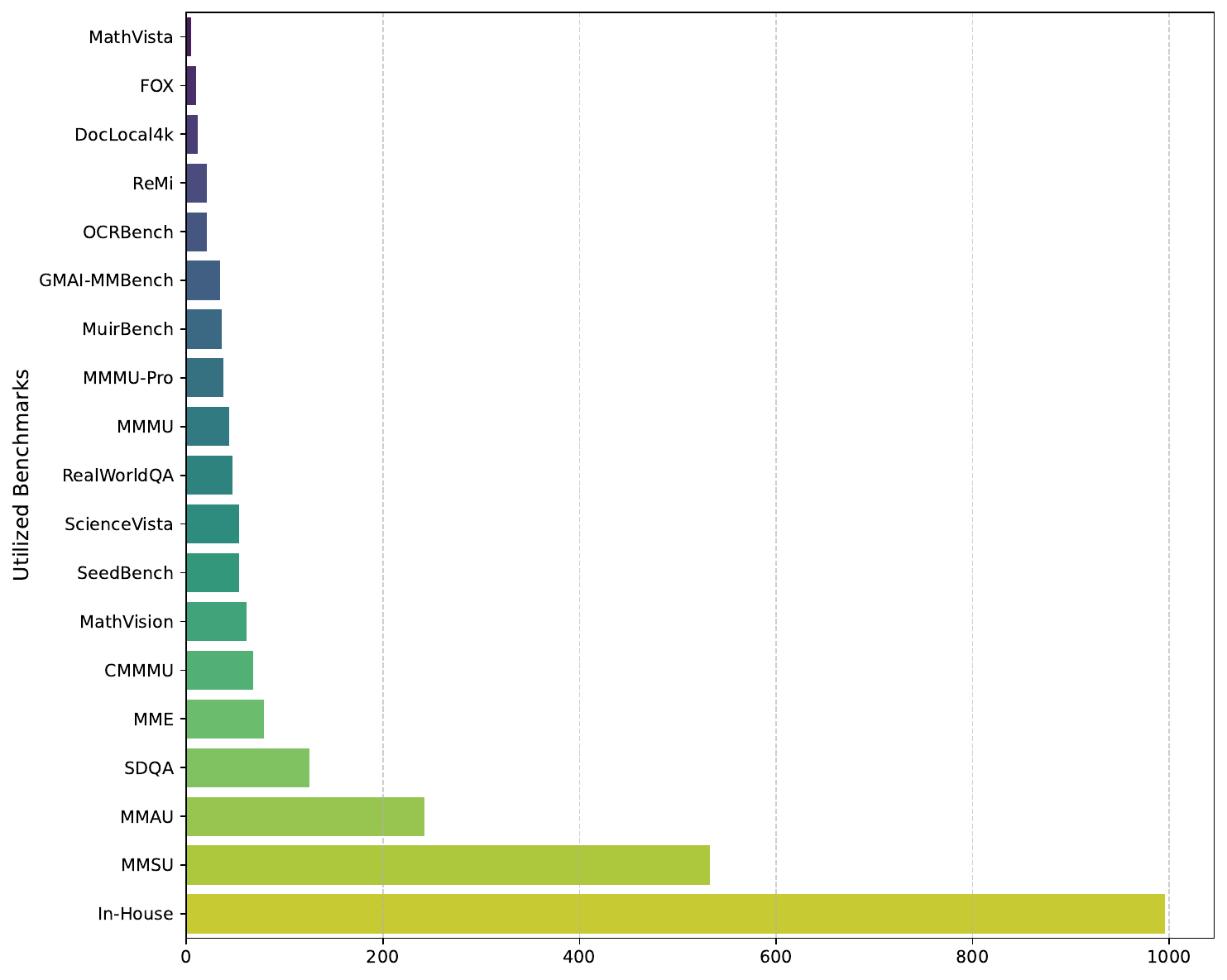}
    \caption{The distribution of the uni-modal benchmarks in UNO-Bench. In addition to publicly available benchmarks, we incorporated several in-house benchmarks both before and after compression to ensure the reasonableness of the data distribution.}
    \label{fig:compresion_source_benchmarks}
\end{figure}

To construct our compressed datasets, we utilized a variety of benchmarks for both visual and audio modalities. For the visual component, we curated data from 15 public and several in-house benchmarks that assess a range of capabilities, including general visual question answering, document and chart comprehension, STEM/scientific reasoning, and multi-image understanding. For the audio component, we used 3 audio question answering benchmarks. The detailed composition of the resulting uni-modal dataset is presented in Figure.\ref{fig:compresion_source_benchmarks}.

\begin{itemize}
    \item \textbf{General visual question answering}, RealWorldQA\cite{realworldqa}, MME\cite{chaoyou2023mme}, SeedBench\cite{li2023seed}.
    \item \textbf{Document and chart understanding}, OCRBench \cite{liu2024ocrbench}, Fox\cite{liu2024focus}, DocLocal4k\cite{ye2023mplug}.
    \item \textbf{Stem \& reasoning}, MMMU\cite{yue2023mmmu}, MMMU-Pro\cite{yue2025mmmupro}, CMMMU\cite{zhang2024cmmmu}, MathVista\cite{lu2024mathvista}, MathVision\cite{wang2024mathvision}, ScienceVista\cite{team2025kimi}, GMAI-MMBench\cite{ye2024gmai}.
    \item \textbf{Multi-image Understanding}, ReMi\cite{kazemi2024remi},  MuirBench\cite{wang2024muirbench}.
    \item \textbf{Audio question answering}, MMAU\cite{mmau}, MMSU\cite{wang2025mmsu}, SDQA\cite{faisal-etal-21-sdqa}.
\end{itemize}

\section{Rigorous Derivation of the Compositional Law}
\label{sec:appendix_derivation_of_law}
Defining the \textbf{performance gain} as $\mathcal{P}'_{\text{Omni}} = \mathcal{P}_{\text{Omni}} - b$. From Eq.~\ref{eq:general_form}, we have:
\begin{equation}
    \mathcal{P}'_{\text{Omni}}(\mathcal{P}_{\text{A}}, \mathcal{P}_{\text{V}}) = f_{\text{A}}(\mathcal{P}_{\text{A}}) + f_{\text{V}}(\mathcal{P}_{\text{V}}) + f_{\text{I}}(\mathcal{P}_{\text{A}}, \mathcal{P}_{\text{V}})
    \label{eq:gain_form}
\end{equation}

Due to the high quality of our benchmark, where a task is unsolvable if either modality is absent, causing the performance to drop to its baseline (e.g. random guessing). we can have a strict boundary condition:
\begin{equation}
\begin{aligned}
    \mathcal{P'}_{\text{Omni}}(\mathcal{P}_{\text{A}}, 0) = 0 \quad &\text{and} \\
    \mathcal{P'}_{\text{Omni}}(0, \mathcal{P}_{\text{V}}) = 0 \quad &\text{and} \\
    \mathcal{P'}_{\text{Omni}}(0,0) = 0
\end{aligned}
\label{eq:boundary_condition}
\end{equation}

Applying the boundary condition of Eq.~\ref{eq:boundary_condition} to Eq.~\ref{eq:gain_form}, we find that the gain is a second-order mixed difference of $f_{\text{I}}$:
\begin{equation}
    \begin{split}
        \mathcal{P}'_{\text{Omni}}(\mathcal{P}_{\text{A}}, \mathcal{P}_{\text{V}}) = & f_{\text{I}}(\mathcal{P}_{\text{A}}, \mathcal{P}_{\text{V}}) - f_{\text{I}}(\mathcal{P}_{\text{A}}, 0) \\
        & - f_{\text{I}}(0, \mathcal{P}_{\text{V}}) + f_{\text{I}}(0, 0)
    \end{split}
    \label{eq:gain_identity}
\end{equation}

Substituting the Taylor series of $f_{\text{I}}$ around $(0,0)$ into Eq.~\ref{eq:gain_identity}, the performance gain is thus exactly equal to the sum of all pure interaction terms from $f_{\text{I}}$:
\begin{equation}
    \begin{split}
        \mathcal{P}'_{\text{Omni}}(\mathcal{P}_{\text{A}}, \mathcal{P}_{\text{V}}) &= \sum_{i \ge 1, j \ge 1} c_{ij} \mathcal{P}_{\text{A}}^i \mathcal{P}_{\text{V}}^j 
    \end{split}
    \label{eq:main_interaction_sum}
\end{equation}
where the coefficients $c_{ij}$ are constants derived from the partial derivatives of $f_{\text{I}}$ at the origin.
For sufficiently small uni-modal performances, we can approximate this series by its leading-order term:
\begin{equation}
    \mathcal{P}'_{\text{Omni}}(\mathcal{P}_{\text{A}}, \mathcal{P}_{\text{V}}) \approx c_{11} \mathcal{P}_{\text{A}} \mathcal{P}_{\text{V}}
\end{equation}
This result strongly motivates modeling the interaction with the general multiplicative Cobb-Douglas form. Re-introducing the baseline $b$ yields our final Compositional Law:
\begin{equation}
    \mathcal{P}_{\text{Omni}} = C \cdot \mathcal{P}_{\text{A}}^{\alpha} \mathcal{P}_{\text{V}}^{\beta} + b
    \label{eq:compositional_law_cobb_1}
\end{equation}

\section{Model Selection for the Compositional Law}
\label{appendix:model_selection}

To validate our choice of the Compositional Law, we compared its performance against several alternative models. The fitting results for all candidate models on our 9-model dataset are summarized in Table.\ref{tab:model_comparison}.

\begin{table*}[t!]
\centering
\caption{Fitting results for all candidate models. While more complex models achieve higher fitting scores ($R^2$), their parameters lack physical interpretability (e.g., negative exponents), indicating severe overfitting on our small dataset. Our chosen \textbf{Symmetric Power Law} offers the best balance of a high $R^2$ value and theoretical soundness.}
\label{tab:model_comparison}
\resizebox{\textwidth}{!}{%
\begin{tabular}{l|cc|l}
\toprule
\textbf{Model Name} & \textbf{$R^2$} & \textbf{RMSE} & \textbf{Fitted Equation} \\
\midrule
Generalized Power Law & \textbf{0.999} & \textbf{0.005} & $P_{Omni} \approx 1.33 \cdot P_{Audio}^{-1.59} \cdot P_{Visual}^{5.09} + 0.24$ \\
Linear Interaction & 0.995 & 0.010 & $P_{Omni} \approx 0.97 - 2.01 P_{Audio} - 0.59 P_{Visual} + 2.85 (P_{Audio} \times P_{Visual})$ \\
Weighted Sum Power Law & 0.995 & 0.010 & $P_{Omni} \approx 1.19 \cdot (-0.20 P_{Audio} + 1.20 P_{Visual})^{3.83} + 0.24$ \\
\textbf{Symmetric Power Law} & \textbf{0.976} & \textbf{0.022} & $\boldsymbol{P_{Omni} \approx 1.03 \cdot (P_{Audio} \times P_{Visual})^{2.19} + 0.24}$ \\
Simple Linear & 0.945 & 0.033 & $P_{Omni} \approx -0.15 - 0.37 P_{Audio} + 1.43 P_{Visual}$ \\
\bottomrule
\end{tabular}%
}
\end{table*}

As shown in Table.\ref{tab:model_comparison}, more complex models like the `Generalized Power Law' achieve a near-perfect fit on the training data. However, this superior performance is misleading. These models yield parameters that are physically implausible, such as negative exponents (e.g., $P_{Audio}^{-1.59}$) or negative weights. Such parameters would illogically imply that improving a model's uni-modal capability could degrade its omni-modal performance. This is a classic symptom of overfitting, where a model with high capacity learns the noise in a small dataset rather than a generalizable underlying trend.

In contrast, our proposed \textbf{Symmetric Power Law} provides an excellent fit ($R^2=0.976$) while maintaining theoretical coherence. All its parameters are positive and have clear interpretations: a super-linear synergy ($\alpha=2.19 > 1$) between modalities, a positive scaling factor ($C=1.03$), and a reasonable baseline score ($b=0.24$). Following the principle of Occam's Razor, we select this model as it offers the most parsimonious, robust, and interpretable explanation for the observed phenomenon.

Interestingly, while the parameters from the overfitted models are invalid, they consistently suggest a stronger influence from the visual modality (e.g., the large positive exponent for $P_{Visual}$ in the `Generalized Power Law`). This hints that while our symmetric law captures the primary collaborative effect, the visual component may play a slightly more dominant role, a direction for future investigation.


\section{Model Ability Taxonomy}
\label{sec:appendix_model ability}
This section will provide specific definitions for each ability item and present examples of various task types.

\begin{figure*}
    \centering
    \includegraphics[width=\linewidth]{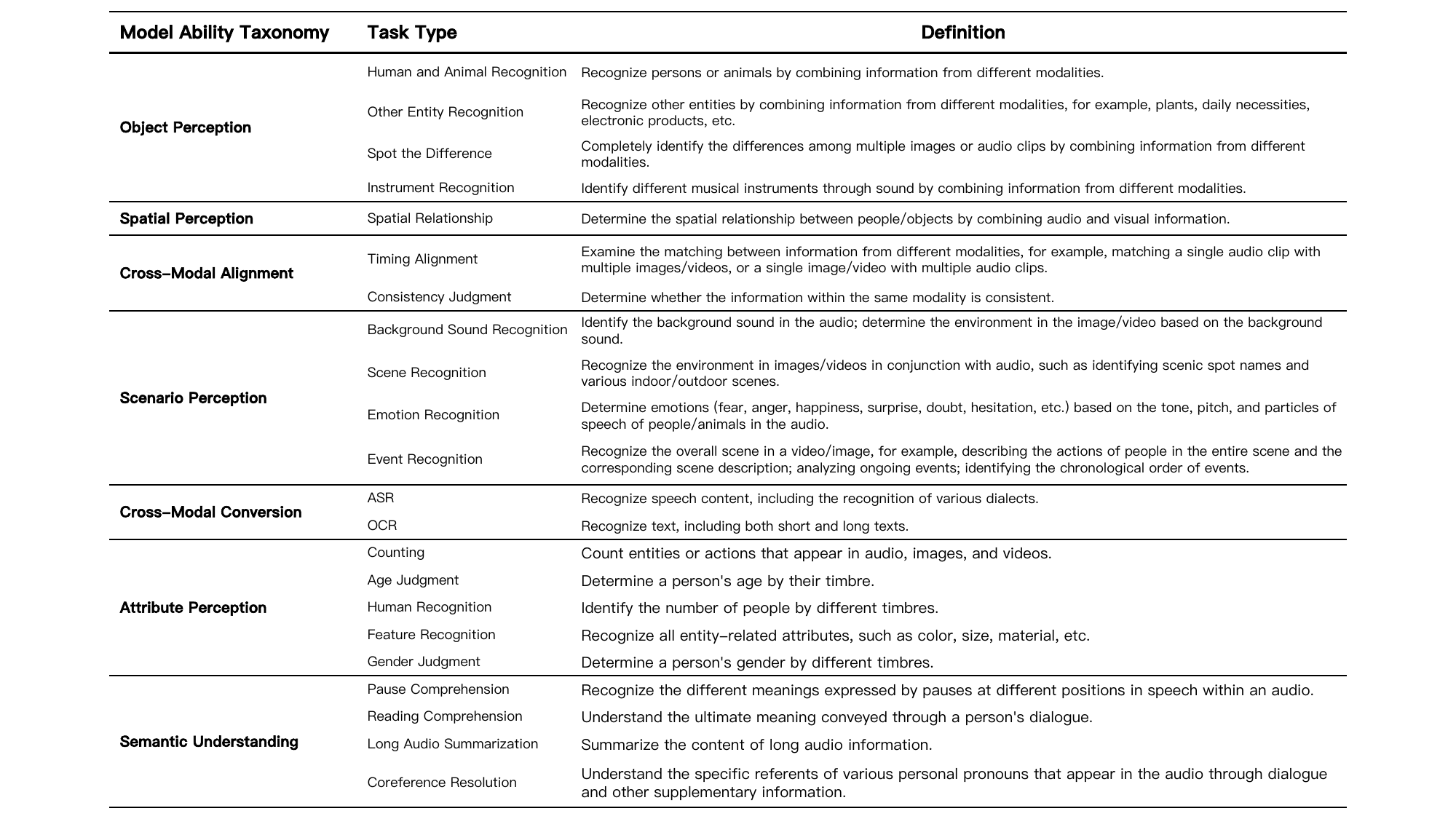}
    \caption{Definition of the Perception Dimension.}
    \label{fig:Definition of the Perception Layer}
\end{figure*}

\begin{figure*}
    \centering
    \includegraphics[width=\linewidth]{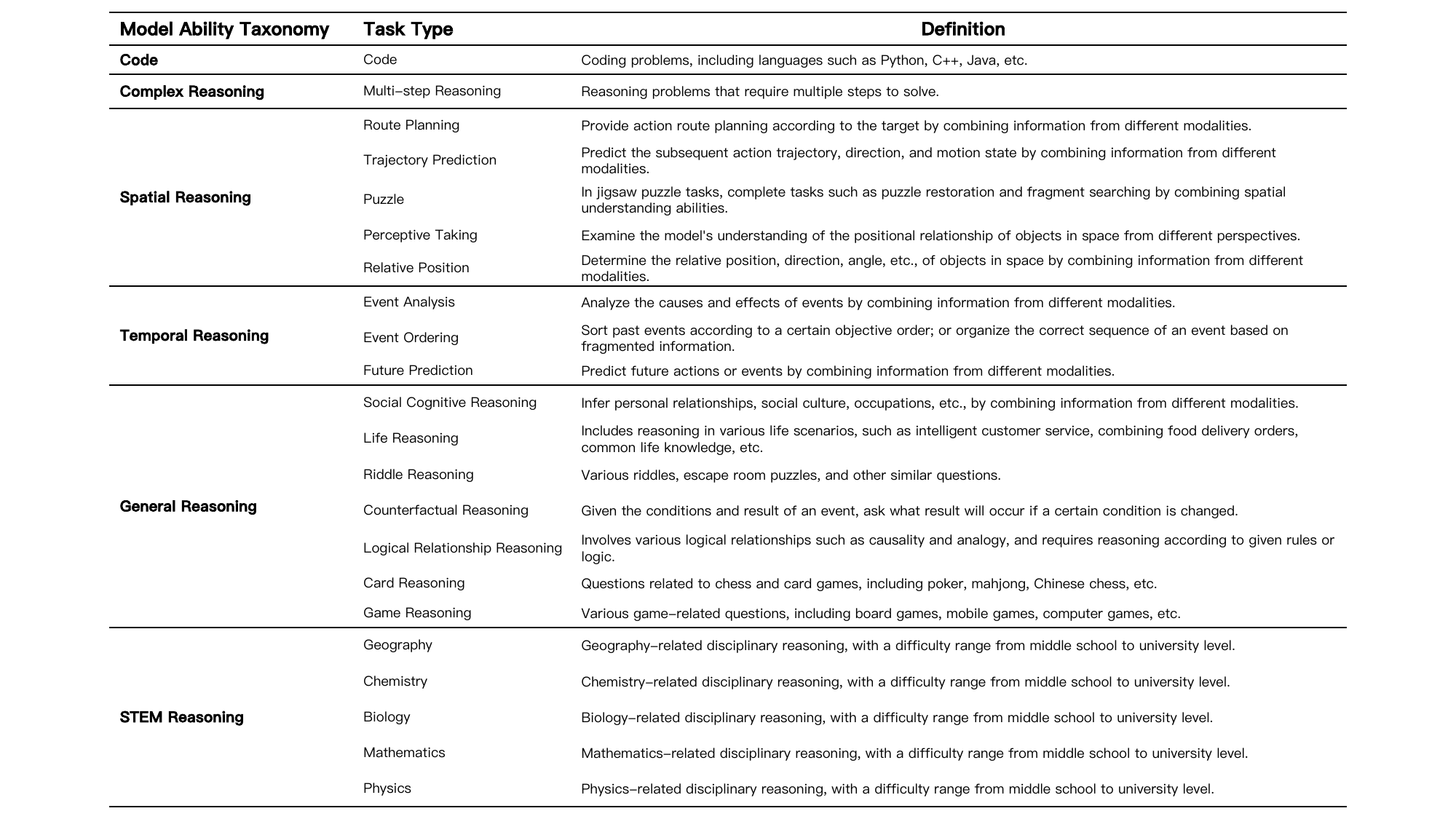}
    \caption{Definition of the Reasoning Dimension.}
    \label{fig:Definition of the Reasoning Layer}
\end{figure*}

\begin{figure*}
    \centering
    \includegraphics[width=\linewidth]{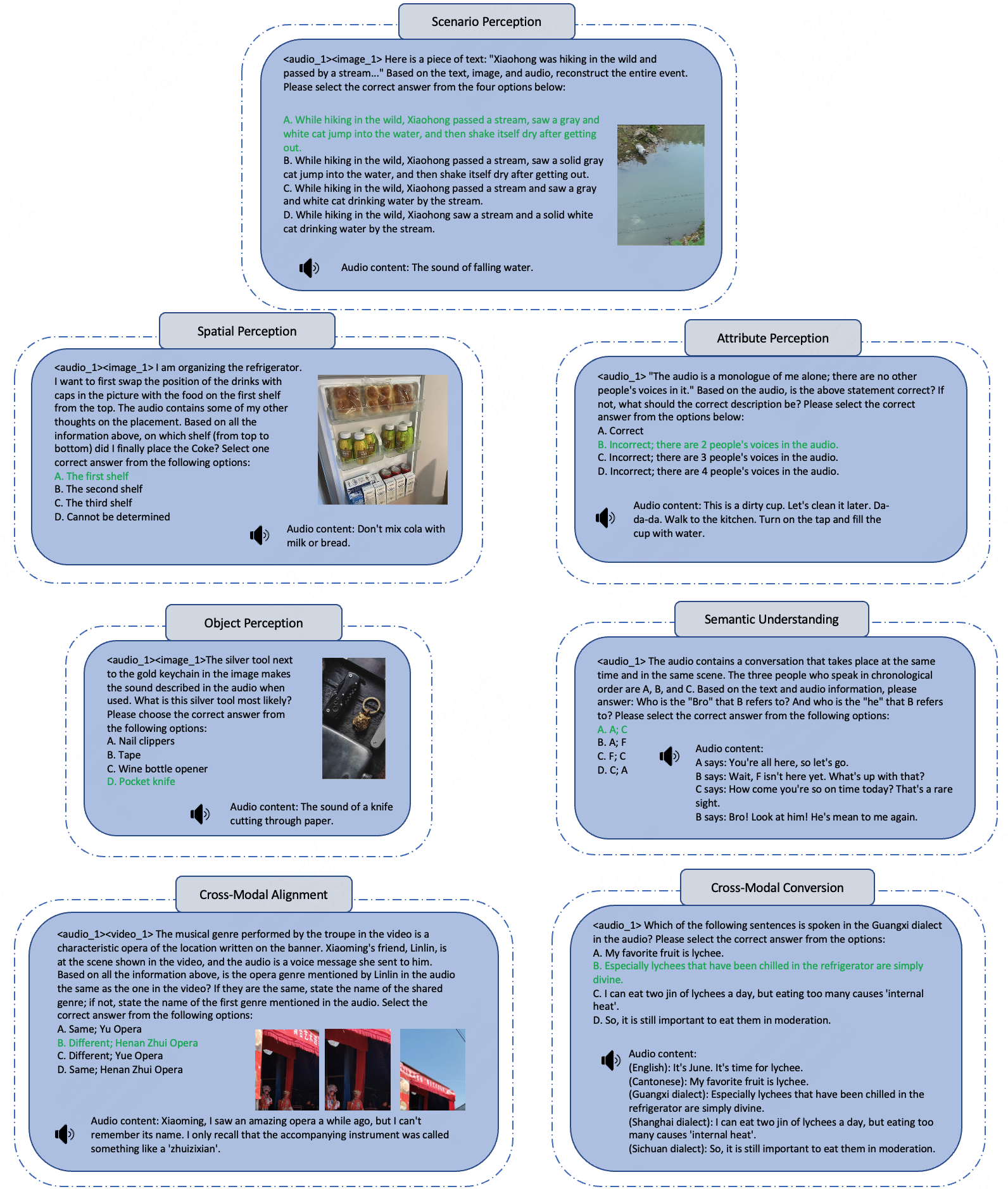}
    \caption{Example of each ability in perception dimension. }
    \label{fig:ability_demo_perception}
\end{figure*}

\begin{figure*}
    \centering
    \includegraphics[width=\linewidth]{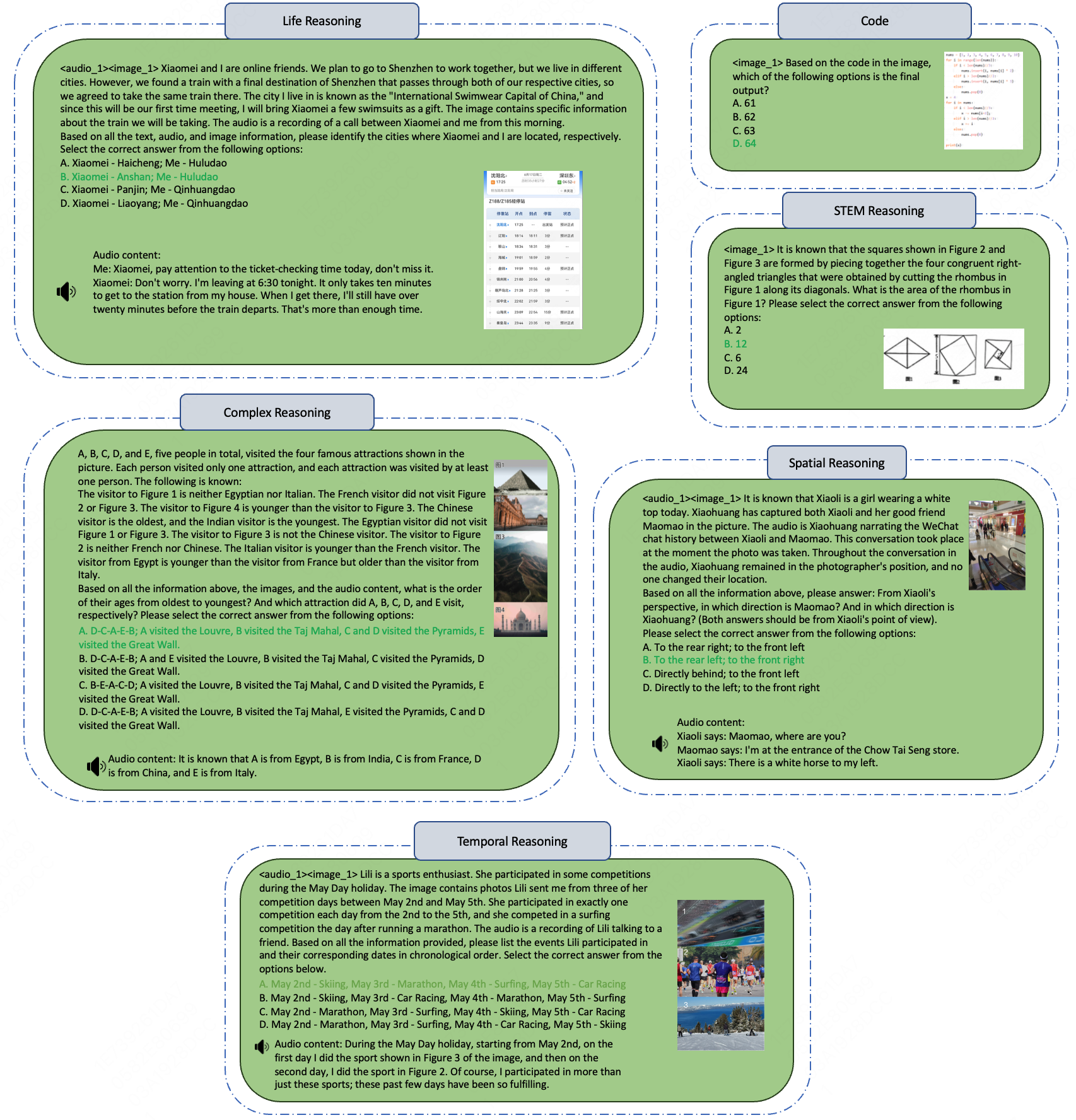}
    \caption{Example of each ability in reason dimension.}
    \label{fig:ability_demo_reason}
\end{figure*}

The specific model abilities and task types of the Perception dimension can be seen in Figure.\ref{fig:Definition of the Perception Layer}, and the Reasoning dimension can be seen in Figure.\ref{fig:Definition of the Reasoning Layer}.
Specific examples are provided for each model ability. Examples of Perception ability including Object Perception, Spatial Perception, Cross-Modal Alignment, Attribute Perception, Scenario Perception, Cross-Modal Conversion and Semantic Understanding can be seen in Figure.\ref{fig:ability_demo_perception}. 
Examples of Reasoning ability including Complex Reasoning, Temporal Reasoning, Spatial Reasoning, Life Reasoning, STEM Reasoning and Code can be respectively seen in Figure.\ref{fig:ability_demo_reason}.

\end{document}